\documentclass[10pt,twocolumn,letterpaper]{article}

\usepackage[pagenumbers]{cvpr} %

\usepackage{graphicx}
\usepackage{amsmath}
\usepackage{amssymb}
\usepackage{booktabs}
\usepackage{color, colortbl}
\usepackage{pifont}
\usepackage{balance}
\usepackage[toc,page,header]{appendix}

\usepackage{booktabs}
\usepackage{balance}
\usepackage{multirow}
\usepackage{cite}
\usepackage[font=small,labelfont=bf]{caption}
\usepackage{color}
\usepackage{xcolor}
\usepackage{tabularx}
\usepackage{arydshln}
\usepackage{amssymb}%
\usepackage{pifont}%

 \usepackage{amsmath}
\usepackage{MnSymbol}%
\usepackage{wasysym}%
\usepackage{enumitem}
\usepackage{wrapfig}
\usepackage{xspace}
\usepackage{tabu}
\usepackage[super]{nth}
\usepackage{scalefnt}
\usepackage{dsfont}
\usepackage{algorithm}
\usepackage{algpseudocode}

\definecolor{darkgreen}{RGB}{0,255,0}
\definecolor{linkgreen}{RGB}{52,130,48}
\definecolor{LightCyan}{rgb}{0.87,0.92,0.96}
\definecolor{m_green}{RGB}{233, 254, 187}
\definecolor{m_orange}{RGB}{255, 212, 121}
\definecolor{m_red}{RGB}{255, 190, 188}
\definecolor{m_violet}{RGB}{215, 131, 255}
\definecolor{m_blue}{RGB}{186, 234, 255}
\definecolor{notetext}{rgb}{0.7,0,0}

\def\eg{\emph{e.g.}\@\xspace} 
\def\ie{\emph{i.e.}\@\xspace} 
 
\def\etc{\emph{etc.}\@\xspace} \def\vs{\emph{vs.}\@\xspace}

\newcommand{\circlenum}[1]{{\textcircled{\scriptsize{#1}}}}
\newcommand{\parag}[1]{\vspace{-5px} \vskip8pt \noindent \textbf{#1}}

\newcommand{\paragt}[1]{\vspace{-5px}\paragraph{#1}\vspace{-5pt}}

\newcommand{\cmark}{\ding{51}}%
\newcommand{\xmark}{\ding{55}}%

\newcommand{\name}{RoomFormer}
\newcommand{\npoly}{M}
\newcommand{\nvert}{N}
\newcommand{\fp}{floorplan}
\newcommand{\fps}{floorplans}
\newcommand{\Fp}{Floorplan}
\newcommand{\gt}{groundtruth}

\newif\ifmynotes
\mynotestrue

\newcommand{\first}[1]{{\textcolor{cyan}{#1}}}
\newcommand{\second}[1]{{\textcolor{orange}{#1}}}

\newcolumntype{Y}{>{\centering\arraybackslash}X}
\newcolumntype{Z}{>{\raggedleft\arraybackslash}X}

\newcommand{\symfootnote}[1]{%
\let\oldthefootnote=\thefootnote%
\stepcounter{mpfootnote}%
\addtocounter{footnote}{-1}%
\renewcommand{\thefootnote}{\fnsymbol{mpfootnote}}%
\footnote{#1}%
\let\thefootnote=\oldthefootnote%
}

\usepackage{array}
\newcolumntype{P}[1]{>{\centering\arraybackslash}p{#1}}
\newcolumntype{M}[1]{>{\centering\arraybackslash}m{#1}}

\usepackage[accsupp]{axessibility}  %

\usepackage[pagebackref,breaklinks,colorlinks]{hyperref}

\usepackage[capitalize]{cleveref}
\crefname{section}{Sec.}{Secs.}
\Crefname{section}{Section}{Sections}
\Crefname{table}{Table}{Tables}
\crefname{table}{Tab.}{Tabs.}

\def\cvprPaperID{1010} %
\def\confName{CVPR}
\def\confYear{2023}

\begin{document}

\title{Connecting the Dots: Floorplan Reconstruction Using Two-Level Queries}

\author{Yuanwen Yue\textsuperscript{1} \quad
Theodora Kontogianni\textsuperscript{2}  
\quad
Konrad Schindler\textsuperscript{1,2} \quad
Francis Engelmann\textsuperscript{2}
\vspace{5px}
\\
\textsuperscript{1}{\normalsize Photogrammetry and Remote Sensing, ETH Zurich}
\quad
\textsuperscript{2}{\normalsize ETH AI Center, ETH Zurich}
\\
\vspace{-15px}
}
\maketitle

\begin{abstract}
We address 2D \fp{} reconstruction from 3D scans. Existing approaches typically employ heuristically designed multi-stage pipelines. Instead, we formulate \fp{} reconstruction as a single-stage structured prediction task: find a variable-size set of polygons, which in turn are variable-length sequences of ordered vertices. To solve it we develop a novel Transformer architecture that generates polygons of multiple rooms in parallel, in a holistic manner without hand-crafted intermediate stages.
The model features two-level queries for polygons and corners, and includes polygon matching to make the network end-to-end trainable. Our method achieves a new state-of-the-art for two challenging datasets, Structured3D and SceneCAD,
along with significantly faster inference than previous methods.
Moreover, it can readily be extended to predict additional information, i.e., semantic room types and architectural elements like doors and windows.
Our code and models are available at: \href{https://github.com/ywyue/RoomFormer}{https://github.com/ywyue/RoomFormer}.
\end{abstract}
\vspace{-5px}

\section{Introduction}
\label{sec:intro}

The goal of \fp{} reconstruction is to turn observations of an (indoor) scene into a 2D vector map in birds-eye view.
More specifically, we aim to abstract a 3D point cloud into a set of closed polygons corresponding to rooms, optionally enriched with further structural and semantic elements like doors, windows and room type labels.

Floorplans are an essential representation that enables a wide range of applications in robotics, AR/VR, interior design, \etc
Like prior work~\cite{stekovic2021montefloor, Chen19ICCV, Chen22CVPR, bassier2020unsupervised, avetisyan2020scenecad}, we start from a 3D point cloud, which can easily be captured with RGB-D cameras, laser scanners or SfM systems.
Several works~\cite{liu2018floornet,Chen19ICCV,stekovic2021montefloor,Chen22CVPR} have shown the effectiveness of projecting the raw 3D point data along the gravity axis, to obtain a 2D density map that highlights the building's structural elements (\eg{}, walls).
We also employ this early transition to 2D image space.
The resulting density maps are compact and computationally efficient, but inherit the noise and data gaps of the underlying point clouds, hence \fp{} reconstruction remains a challenging task.

\begin{figure}[t]
    \centering
    \begin{tabular}{ccc}
    {\small \em Input 3D Point Cloud} & \hspace{12px} & {\small  \em Reconstructed Floorplan} \\
    \end{tabular}
    \includegraphics[width=0.45\textwidth]{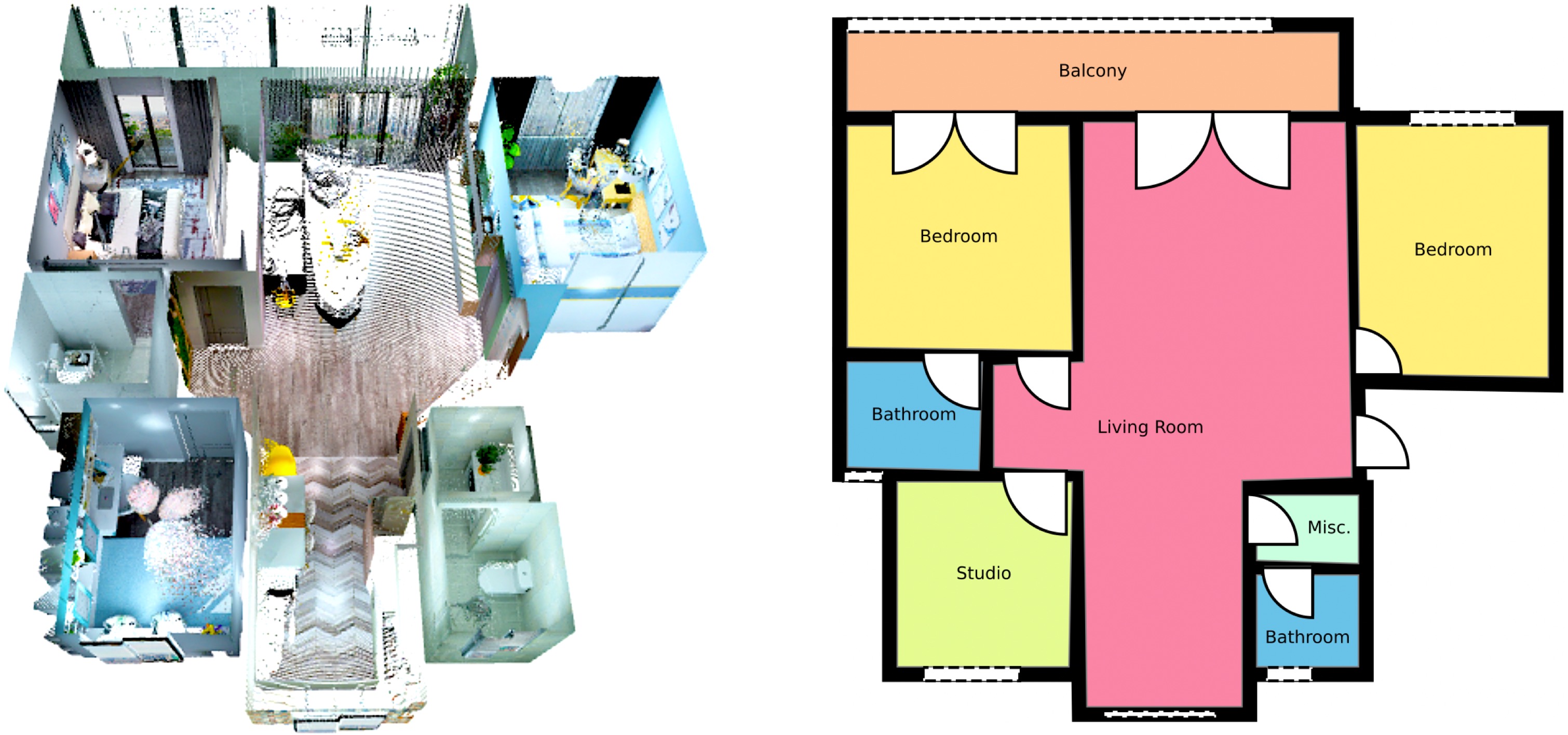}  %
    \caption{\textbf{Semantic floorplan reconstruction.}
    Given a point cloud of an indoor environment, \emph{\name{}} jointly recovers multiple room polygons along with their associated room types, as well as architectural elements such as doors and windows.}
    \label{fig:teaser}
    \vspace{-16px}
\end{figure}

Existing methods can  be split broadly into two categories that both operate in two stages:
\emph{Top-down methods}~\cite{Chen19ICCV,stekovic2021montefloor} first extract room masks from the density map using neural networks
(\eg{}, Mask R-CNN~\cite{he2017mask}), then employ optimization/search techniques
(\eg{}, integer programming~\cite{schrijver1998theory},
Monte-Carlo Tree-Search~\cite{browne2012survey}) to extract a polygonal \fp{}.
Such techniques are not end-to-end trainable, and their success depends on how well the hand-crafted optimization captures domain knowledge about room shape and layout.
Alternatively, \emph{bottom-up methods}~\cite{liu2018floornet,Chen22CVPR} first detect corners, then look for edges between corners (\ie{}, wall segments)
and finally assemble them into a planar \fp{} graph.
Both approaches are strictly sequential and therefore dependent on the quality of the initial corner, respectively room, detector. The second stage starts from the detected entities, therefore missing or spurious detections may significantly impact the reconstruction.

We address those limitations and design a model that directly maps a density image to a set of room polygons.
Our model, named \emph{\name{}}, leverages the sequence prediction capabilities of Transformers and directly outputs a variable-length, ordered sequence of vertices per room. \name{} requires neither hand-crafted, domain-specific intermediate products nor explicit corner, wall or room detections.
Moreover, it predicts all rooms that make up the \fp{} at once, exploiting the parallel nature of the Transformer architecture.

In more detail, we employ a standard CNN backbone to extract features from the birds-eye view density map, followed by a Transformer encoder-decoder setup that consumes image features (supplemented with positional encodings) and outputs multiple ordered corner sequences, in parallel.
The \fp{} is recovered by simply connecting those corners in the predicted order.
Note that the described process relies on the ability to generate hierarchically structured output of variable and a-priori unknown size, where each \fp{} has a different number of rooms (with no natural order), and each room polygon has a different number of (ordered) corners. 
We address this challenge by introducing two-level queries with one level for the room polygons and one level for their corners. The varying numbers of both rooms and corners are accommodated by additionally classifying each query as valid or invalid.
The decoder iteratively refines the queries, through self-attention among queries and cross-attention between queries and image features.
To enable end-to-end training, we propose a polygon matching strategy that establishes the correspondence between predictions and  targets, at both room and corner levels.
In this manner, we obtain an integrated model that holistically predicts a set of polygons to best explain the evidence in the density map, without hand-tuned intermediate rules of 
which corners, walls or rooms to commit to along the way.
The model is also fast at inference, since it operates in single-stage feed-forward mode, without optimization or search and without any post-processing steps.
Moreover, it is flexible and can, with few straight-forward modifications, predict additional semantic and structural information such as room types, doors and windows (Fig.\,\ref{fig:teaser}).

We evaluate our model on two challenging datasets, Structured3D~\cite{zheng2020structured3d} and SceneCAD~\cite{avetisyan2020scenecad}.
For both of them, \name{} outperforms the state of the art, while at the same time being significantly faster than existing methods.
In summary, our contributions are:
\begin{itemize}[noitemsep,nolistsep]
\vspace{-1px}
\item A new formulation of \fp{} reconstruction, as the simultaneous generation of multiple ordered sequences of room corners. 
\item The \name{} model, an end-to-end trainable, Transformer-type architecture that implements the proposed formulation via two-level queries that predict a set of polygons each consisting of a sequence of vertex coordinates.
\item Improved \fp{} reconstruction scores on both Structured3D~\cite{zheng2020structured3d} and SceneCAD~\cite{avetisyan2020scenecad}, with faster inference times.
\item Model variants able to additionally predict semantic room type labels, doors and windows.
\end{itemize}

\section{Related Work}
\label{sec:related_work}
\parag{Floorplan reconstruction}
turns raw sensor data (\eg{}, point clouds, density maps, RGB images) into vectorized geometries.
Early methods rely on basic image processing techniques, \eg, Hough transform or plane fitting~\cite{llados1997system,okorn2010toward, adan20113d,budroni2010automated,sanchez2012planar,xiong2013automatic,monszpart2015rapter}.
Graph-based methods~\cite{furukawa2009reconstructing,cabral2014piecewise,ikehata2015structured} cast \fp{} reconstruction as an energy minimization problem. 
Recent deep learning methods replace some hand-crafted components with neural networks.
Typical top-down methods such as
Floor-SP\cite{Chen19ICCV} rely on Mask R-CNN\cite{he2017mask} to detect room segments and reconstruct polygons of individual room segments by sequentially solving shortest path problems.
Similarly, MonteFloor~\cite{stekovic2021montefloor} first detects room segments and then relies on Monte-Carlo Tree-Search to select room proposals. %
Alternative bottom-up methods, such as
FloorNet~\cite{lin2019floorplan} first detect room corners, followed by an integer programming formulation to generate wall segments. %
This approach, however, is limited to Manhattan scenes.
Recently, HEAT~\cite{Chen22CVPR} proposed an end-to-end model following a 
typical bottom-up pipeline:
first detect corners, then classify edge candidates between corners.
Although end-to-end trainable, it cannot recover edges from undetected corners. %
Instead, our approach skips the heuristics-guided processes from both approaches. %
Without explicit corner, wall or room detection,
our approach directly generates rooms as polygons in a holistic fashion.
\paragt{Transformers for structured reconstruction.}
Transformers~\cite{vaswani2017attention}, originally proposed for sequence-to-sequence translation tasks,
have shown promising performance in many vision tasks such as object detection~\cite{carion2020end, zhu2020deformable}, image/video segmentation~\cite{xie2021segformer,cheng2021per, cheng2021mask2former} and tracking~\cite{meinhardt2022trackformer}. 
DETR~\cite{carion2020end} reformulates object detection as a direct set prediction problem with Transformers which is free from many hand-crafted components, \eg{}, anchor generation and non-maximum suppression.
LETR~\cite{xu2021line} extends DETR by adopting Transformers to predict a set of line segments. 
PlaneTR~\cite{tan2021planetr} follows a similar paradigm 
for plane detection and reconstruction.
These works show the promising potential of Transformers for structured reconstruction without heuristic designs.
Our work goes beyond these initial steps and asks the question:
\emph{Can we leverage Transformers for structured polygon generation?}
Different from predicting primitive shapes that can be represented by a fixed number of parameters~(\eg{}, bounding boxes, lines, planes), polygons are more challenging due to the arbitrary number of (ordered) vertices.%
While some recent works~\cite{liang2020polytransform,lazarow2022instance,zhang2022text} utilize Transformers for polygon generation in the context of instance segmentation or text spotting, there are two essential differences:
(1) They assume a fixed number of polygon vertices, which is not suitable for floorplans.
This results in over-redundant vertices for simple shapes and insufficient vertices for complex shapes.
Instead, our goal is to generate polygons that match the target shape with the correct number of vertices.
(2) They rely on bounding box detection as instance initialization,
while our single-stage method directly generates multiple polygons in parallel.%

\begin{figure*}
    \centering
    \includegraphics[width=1\textwidth]{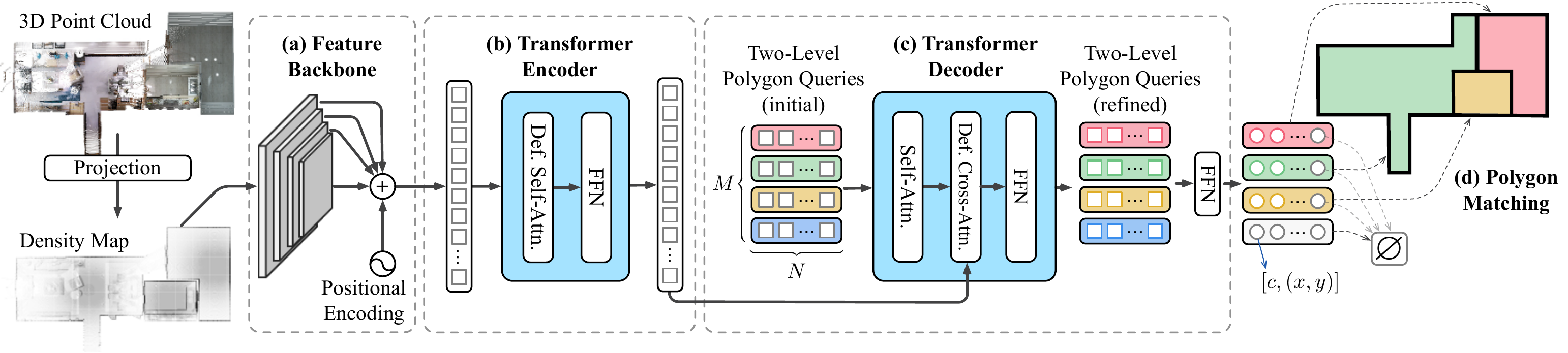}
    \caption{
    \textbf{Illustration of the \name{} model.}
    Given a top-down-view density map of the input point cloud,
    (a) the feature backbone extracts multi-scale features, adds positional encodings, and flattens them before passing them into the (b) Transformer encoder.
    (c) The Transformer decoder takes as input our \emph{two-level} queries, one level for the room polygons (up to $M$) and one level for their corners (up to $N$ per room polygon).
    A feed-forward network (FFN) predicts a class $c$ for each query to accommodate varying numbers of rooms and corners.
    During training, the polygon matching guarantees optimal assignment between predicted and \gt{} polygons.
    }
    \vspace{-10px}
    \label{fig:model}
\end{figure*}

\section{Method}
\label{sec:method}

\subsection{\Fp{} Representation}
A suitable floorplan representation is key to an efficient \fp{} reconstruction system.
Intuitively, one can decompose \fp{} reconstruction as intermediate geometric primitives  detection problems (corners, walls, rooms) and tackle them separately,
as in prior works \cite{Chen19ICCV,stekovic2021montefloor,liu2018floornet,Chen22CVPR}.
However, such pipelines involve heuristics-driven designs and lack holistic reasoning capabilities.

Our core idea is to cast \fp{} reconstruction as a direct set prediction problem of polygons.
Each polygon represents a room and is modeled as an ordered sequence of vertices.
The edges (\ie{}, walls) are implicitly encoded by the order of the vertices -- two consecutive vertices are connected --
thus a separate edge prediction step is not required.
Formally, the goal is to predict \emph{a set of sequences} of arbitrary length, defined as $S=\left \{ V_{m} \right \}_{m=1}^{M^{\text{gt}}}$, where $M^{\text{gt}}$ is the number of sequences per scene, and each sequence $V_{m}=\left( v_{1}^{m}, v_{2}^{m}, ..., v_{N_{m}}^{m} \right )$ represents a closed polygon (\ie{}, room) defined by 
$N_{m}$
ordered vertices.

As each polygon has an arbitrary number of vertices\,$N_m$, we model each vertex $v_{n}^{m}$ in a polygon $V_{m}$ by two variables $v_{n}^{m}=\left ( c_{n}^{m}, p_{n}^{m} \right )$, where $c_{n}^{m}\in \left \{ 0, 1 \right \}$ indicates whether $v_{n}^{m}$ is a valid vertex or not, and $p_{n}^{m}\in \mathbb{R}^{2}$ are the 2D coordinates of the corner in the \fp{}.
Once the model predicts the ordered corner sequences, we connect all valid corners to obtain the polygonal representation of all rooms.

\subsection{Architecture Overview}\label{sec:overview}

Fig.\,\ref{fig:model} shows the model architecture.
It consists of (a) a feature backbone that extracts image features,
(b) a Transformer encoder to refine the CNN features,
and (c) a Transformer decoder using \emph{two-level} queries for polygon prediction.\,(d) During training, the polygon matching module yields optimal assignments between predicted and \gt{} polygons, enabling end-to-end supervision.

\parag{CNN backbone.}
The backbone extracts pixel-level feature maps from the density map $\mathbf{x}_d \in \mathbb{R}^{H \times W} $.
Since both local and global contexts are required for accurately locating corners and capturing their order, we utilize the
$L$  multi-scale feature maps $\left \{ I_{l} \right \}_{l=1}^{L}$ from each layer $l$ of the convolutional backbone,
where $I_{l}\in \mathbb{R}^{C\times H_{l}\times W_{l}}$.
Each feature map is flattened to a feature sequence $I_{l}\in \mathbb{R}^{C\times H_{l}W_{l}}$ and sine/cosine positional encodings $E_{l}\in \mathbb{R}^{C\times H_{l}W_{l}}$ are added to each pixel location.
The flattened feature maps are concatenated and serve as multi-scale input to the Transformer encoder. 

\parag{Multi-scale deformable attention.}
To avoid the computational and memory complexities of standard Transformers ~\cite{vaswani2017attention},
we adopt deformable attention from \cite{zhu2020deformable}.
Given a feature map, for each query element, the deformable attention only attends to a small set $N_{s}$ of key sampling points around a reference point, instead of looking over all $H_{l}W_{l}$ spatial locations on the feature map, where $N_{s}<<H_{l}W_{l}$.
The multi-scale deformable attention applies deformable attention across multi-scale feature maps and enables encoding richer context.
We use multi-scale deformable attention for the self- and cross-attention in the encoder and decoder.

\parag{Transformer encoder}
takes as input the position-encoded multi-scale feature maps and outputs enhanced feature maps of the same resolution. Each encoder layer consists of a multi-scale deformable self-attention module~(MS-DSA) and a feed forward network~(FFN).
In the MS-DSA module,
both the query and key elements are pixel features from the multi-scale feature maps. The reference point is the coordinate of each query pixel.
A learnable scale-level embedding is added to the feature representation to identify which feature level each query pixel lies in.

\parag{Transformer decoder} is stacked with multiple layers (Fig.\,\ref{fig:decoder}).
Each layer consists of a self-attention module (SA),
a multi-scale deformable cross-attention module (MS-DCA) and an FFN.
Each decoder layer takes in the enhanced image features from the encoder and a set of polygon queries from the previous layer.
The polygon queries first interact with each other in the SA module.
In the MS-DCA, the polygon queries attend to different regions of the density map.
Finally, the output of the decoder is passed to a shared FFN to predict binary class labels $c$ for each query indicating its validity as a corner.

\subsection{Modeling \Fp{} As Two-Level Queries}\label{sec:two_query}
\begin{figure}
     \centering
     \begin{subfigure}[b]{0.23\textwidth}
         \centering
         \includegraphics[width=\textwidth]
        {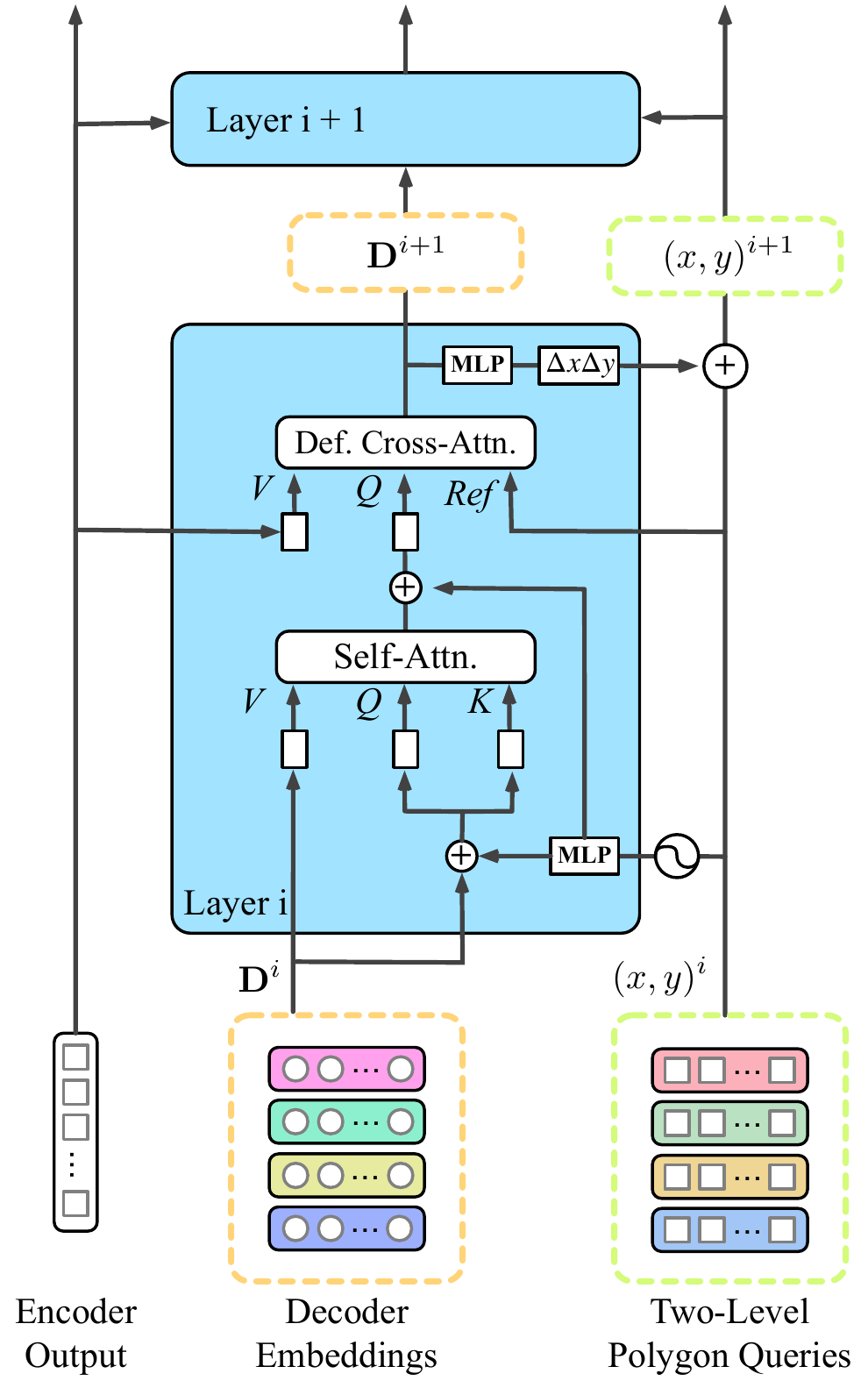}
         \caption{Structure of the decoder}
         \label{fig:decoder}
     \end{subfigure}
     \hfill
     \begin{subfigure}[b]{0.2\textwidth}
         \centering
         \includegraphics[width=\textwidth]{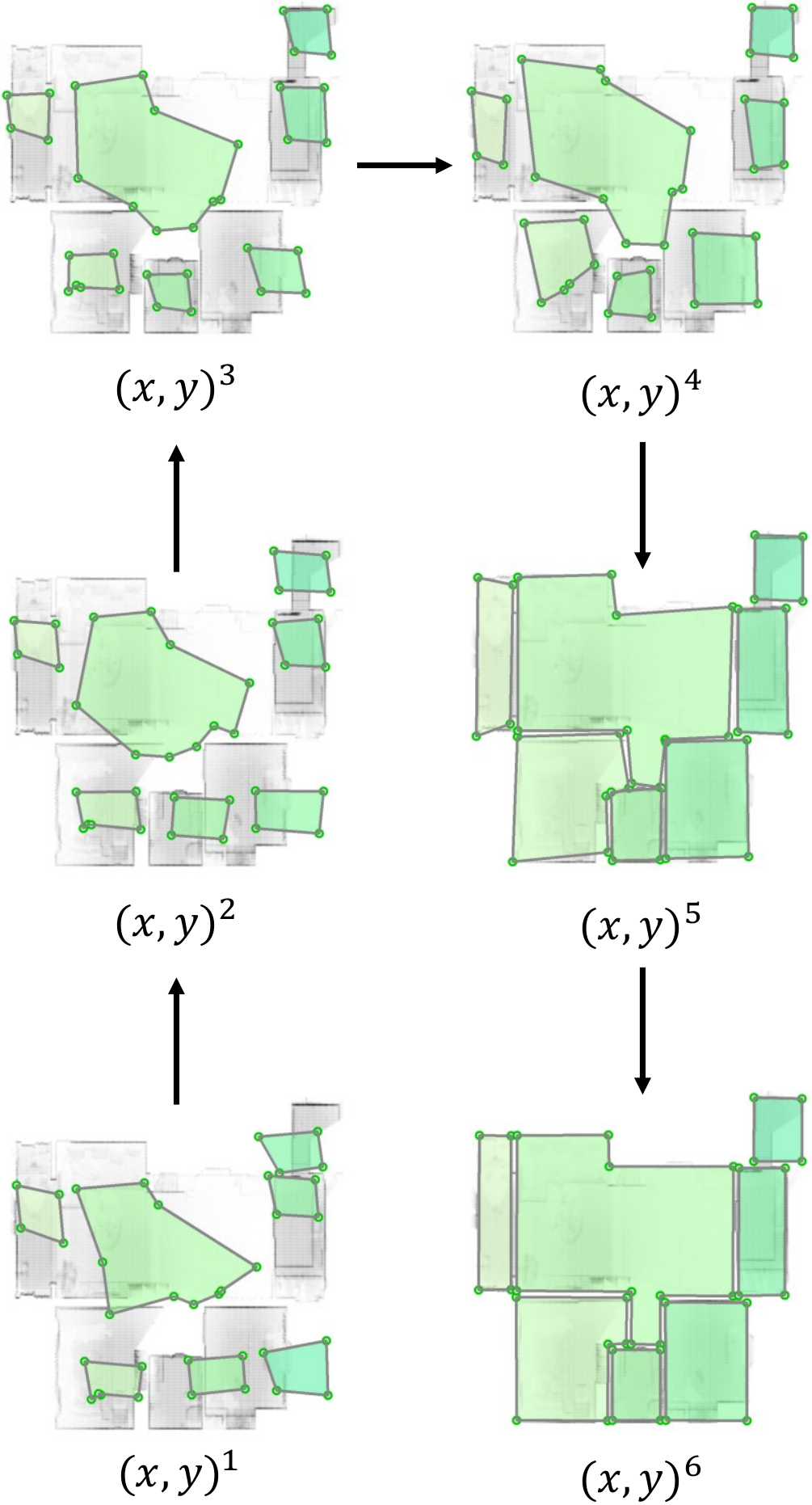}
         \caption{Polygon queries evolution}
         \label{fig:evolve}
     \end{subfigure}
     \caption{(a) Illustration of one layer of the Transformer decoder (we omit the FFN blocks for clarity).
     (b) Visualization of the evolution of the polygon queries after each decoder layer. 
     }
     \label{fig:poly_refine}
         \vspace{-10px}

\end{figure}

We model \fp{} reconstruction as a prediction of a set of sequences.
This motivates the two-level polygon queries, one level for room polygons and one level for their vertices.
Specifically, we represent polygon queries as $Q\in \mathbb{R}^{ \npoly \times \nvert \times 2}$,
where $M$ is the maximum number of polygons (\ie{}, room level),
$N$ is the maximum number of vertices per polygon (\ie{}, corner level).
Using this formulation, we can directly learn \emph{ordered} corner coordinates for each room as queries, which are subsequently refined after each layer in the decoder~(Fig.\,\ref{fig:evolve}).

We illustrate the structure of one decoder layer in Fig.\,\ref{fig:decoder}.
The queries in the decoder consist of two parts: \emph{content} queries (\ie{}, decoder embeddings) and \emph{positional} queries
(generated from polygon queries).
We denote $Q^{i}=(x,y)^{i}$ as the polygon queries in decoder layer $i$\footnote{For simplicity, we drop the polygon and vertex indices.}, and~$D^{i}\in~\mathbb{R}^{ \npoly \times \nvert \times C}$ and $P^{i}\in \mathbb{R}^{ \npoly \times \nvert \times C}$ as the corresponding content and positional queries.
Given the polygon query $Q^{i}$, its positional query $P^{i}$ is generated as
$P^{i}=~\textrm{MLP}(\textrm{PE}(Q^{i}))$, where PE (Positional Encoding) maps the 2D coordinates to a $C$-dimensional sinusoidal positional embedding. The decoder performs self-attention on all corner-level queries regardless of the room they belong to. This simple design not only allows the interaction between corners of a single room,
but also enables interaction among corners across different rooms (\eg{}, corners on the adjacent walls of two rooms are influencing each other),
thus enabling global reasoning.
In the multi-scale deformable attention module, we directly use polygon queries as reference points, allowing us to use explicit spatial priors to pool features from the multi-scale feature maps around the polygon vertices. The varying numbers of both rooms and corners are achieved by classifying each query as valid or invalid. In  Sec.\,\ref{sec:poly_match}, we describe the polygon matching strategy that encourages the queries at corner level to follow a specific order (\ie{}, a sequence) while the queries at room level can be un-ordered (\ie{}, a set).

The key advantage of the above approach is that the room polygons can directly be obtained by connecting the valid vertices in the provided order, without the need for an explicit edge detector as in prior bottom-up methods, \eg{}, \cite{Chen22CVPR}.

\parag{Iterative polygon refinement.}
Inspired by iterative bounding box refinement in~\cite{zhu2020deformable}, we refine the vertices in each polygon in the decoder layer-by-layer.
We use a prediction head (MLP) to predict relative offsets $(\Delta x, \Delta y )$ from the decoder embeddings and update the polygon queries for the next layer.
Both decoder embeddings and polygon queries input to the first layer are initialized from a
normal distribution and learned as part of the model parameters.
During inference, we directly load the learned decoder embeddings and polygon queries and update them layer-by-layer.
We visualize this iterative refinement process in Fig.\,\ref{fig:evolve}.
The final predicted labels are used to select valid queries and visualize their position after each layer.

\begin{figure*}[ht]
     \centering
     \begin{subfigure}[b]{0.3\textwidth}
         \centering
         \includegraphics[width=\textwidth]
        {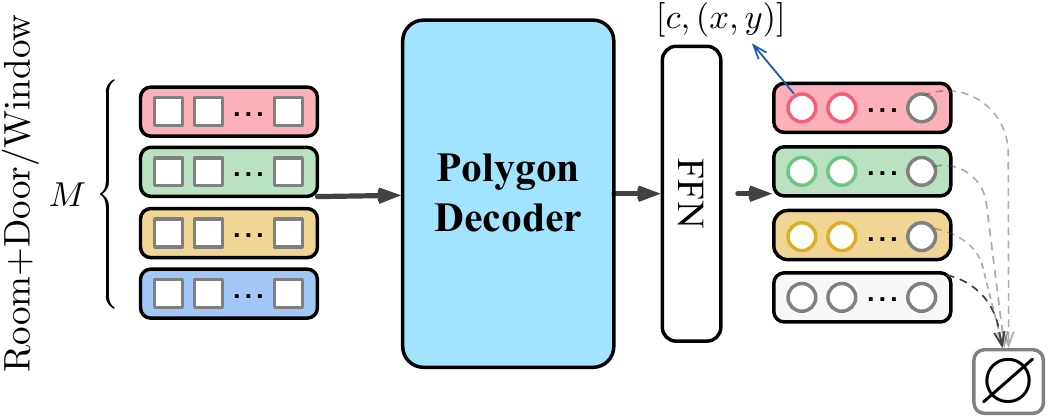}
         \caption{SD-TQ}
         \label{fig:var1}
     \end{subfigure}
     \hfill
     \begin{subfigure}[b]{0.3\textwidth}
         \centering
         \includegraphics[width=\textwidth]{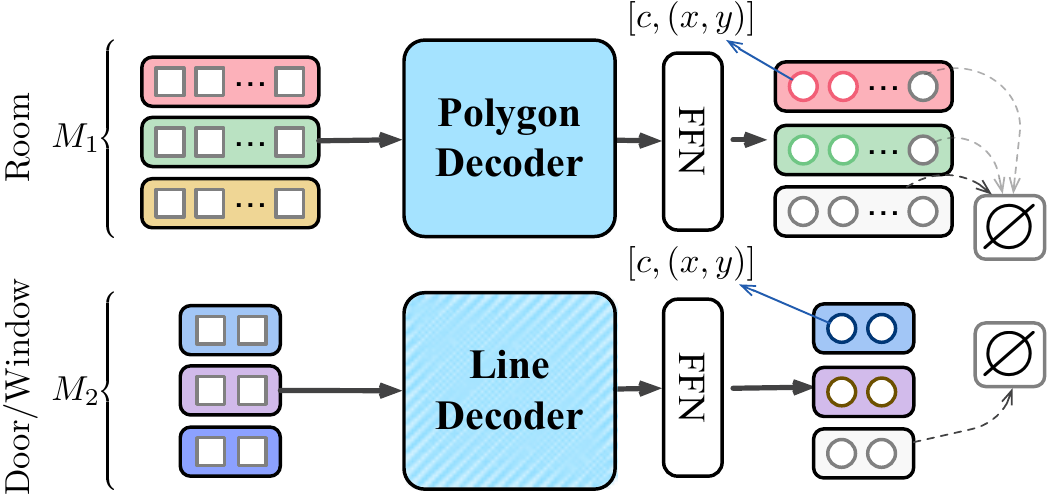}
         \caption{TD-TQ}
         \label{fig:var2}
     \end{subfigure}
     \hfill
     \begin{subfigure}[b]{0.3\textwidth}
         \centering
         \includegraphics[width=\textwidth]{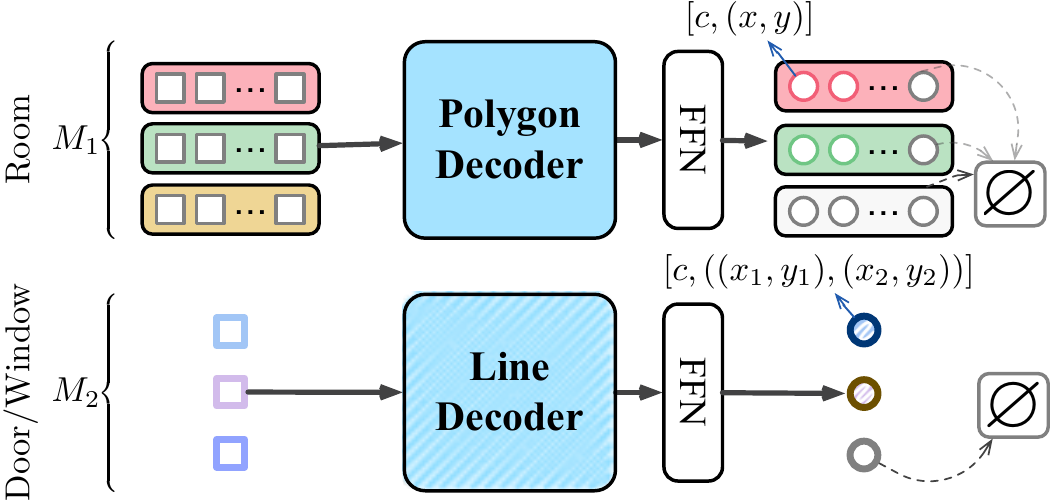}
         \caption{TD-SQ}
         \label{fig:var3}
     \end{subfigure}
     \caption{\textbf{Model variants for semantically-rich \fp{}s.} SD-TQ: Single decoder with two-level queries. TD-TQ: two decoders with two-level queries. TD-SQ: two decoders with single-level queries in the line decoder.
     }
     \label{fig:model_variant_sem_rich}
     \vspace{-10px}
\end{figure*}

\subsection{Polygon Matching}\label{sec:poly_match}
The prediction head of the Transformer decoder outputs a fixed-number $\npoly$ of polygons with a fixed-number $\nvert$ vertices (including non-valid ones, mapped to $\varnothing$) while the \gt{} contains an arbitrary number of polygons with an arbitrary number of vertices.
One of the challenges is to match the fixed-number predictions with the arbitrary-numbered \gt{} to make the network end-to-end trainable.
To this end, we introduce a strategy to handle the matching at two levels: set and sequence level.

Let us denote $\hat{S} = \left \{ \hat{V}_{m}=(\hat{v}_{1}^{m}, \hat{v}_{2}^{m}, ..., \hat{v}_{N}^{m}) \right \}_{m=1}^{M}$ as a set of predicted polygon instances.
Each predicted vertex is represented as $\hat{v}_{n}^{m}=\left ( \hat{c}_{n}^{m}, \hat{p}_{n}^{m} \right )$, where
$\hat{c}_{n}^{m}$ indicates the probability of a valid vertex and $\hat{p}_{n}^{m}$ is the predicted 2D coordinates in normalized space $[0, 1]$. 
Assume there are $M^{\text{gt}}$ polygons in the \gt{} set $S = \left \{ V_{m} \right \}_{m=1}^{M^{\text{gt}}}$, where $V_{m}$ has a length of $N_{m}$. 
For each \gt{} polygon, we first pad it to a length of $N$ vertices so that $V_{m}=(v_{1}^{m}, v_{2}^{m}, ..., v_{N}^{m})$, where $v_{n}^{m}=\left ( c_{n}^{m}, p_{n}^{m} \right )$ and $c_{n}^{m}$ maybe 0, \ie{}, $\varnothing$. Then we further pad $S$ with additional $M-M^{\text{gt}}$ polygons full of $N$ $\varnothing$ vertices so that $\forall (m\in \left [ M^{\text{gt}}+1,M \right ] \wedge  n\in \left [ 1,N \right ]) ,c_{n}^{m}=0$.
At the set level, we find a bipartite matching between the predicted polygons and the \gt{} by searching for a permutation $\hat{\sigma}$ with minimal cost:

\begin{equation}\label{eq:matcher}
    \hat{\sigma} = \underset{\sigma}{\mathrm{arg}\,  \mathrm{min}}\sum_{m=1}^{M}\mathcal{D}(V_{m},\hat{V}_{\sigma(m)})
\end{equation} 
where $\mathcal{D}$ is a function that measures the matching cost between \gt{} polygon $V_{m}$ and a prediction with index $\sigma(m)$. Since we view a polygon as a sequence of vertices, we calculate the matching cost at sequence level and define $\mathcal{D}$ as 
$\mathds{1}_{\left \{ m\leq M^{\text{gt}} \right \}} \lambda_{\mathrm{cls}} \sum_{n=1}^{N}\left \| c_{n}^{m}-\hat{c}_{n}^{\sigma (m)} \right \| + \mathds{1}_{\left \{ m\leq M^{\text{gt}} \right \}} \lambda_{\mathrm{coord}}d(P_{m},\hat{P}_{\sigma (m)})
$,
where $d$ measures the sum of pair-wise $L_{1}$ distance between \gt{} vertex coordinates \emph{without} padding $P_{m}=(p_{1}^{m}, p_{2}^{m}, ..., p_{N_{m}}^{m})$ and the prediction sliced with the same length $\hat{P}_{\sigma (m)}=(\hat{p}_{1}^{\sigma (m)}, \hat{p}_{2}^{\sigma (m)}, ..., \hat{p}_{N_{m}}^{\sigma (m)})$. 
The matching cost $\mathcal{D}$ takes into account both the vertex label and coordinates with balancing coefficients $\lambda_{\mathrm{cls}}$ and $\lambda_{\mathrm{coord}}$.
A closed polygon is a cycle so there exist multiple equivalent parametrizations depending on the starting vertex and the orientation. Here, we fix the \gt{} $P_{m}$ to always follow counter-clockwise orientation, but can start from any of the vertices. We calculate the distance between $\hat{P}_{\sigma (m)}$ and all possible permutations of $P_{m}$ and take the minimum as the final $d$.

\parag{Loss functions.} 
After finding the optimal permutation $\hat{\sigma}$ with the Hungarian algorithm,
we can compute the loss function which consists of three parts:
a vertex label classification loss,
a vertex coordinates regression loss
and a polygon rasterization loss.
The vertex label classification loss is a standard binary cross-entropy:
\begin{equation}
\mathcal{L}_\text{cls}^{m} = -\frac{1}{N}\sum_{n=1}^{N}c_{n}^{m}\cdot \mathrm{log}(\hat{c}_{n}^{\hat{\sigma}(m)})-(1-c_{n}^{m})\cdot  \log(1-\hat{c}_{n}^{\hat{\sigma}(m)})
\end{equation}
Similar to the matching cost function, the $L_1$ distance serves as a loss function for vertex coordinates regression:
\begin{equation}
\mathcal{L}_{\text{coord}}^{m} = \frac{1}{N_{m}} \mathds{1}_{\left \{ m\leq M^\text{gt} \right \}} d(P_{m}, \hat{P}_{\hat{\sigma}(m)})
\end{equation}
We additionally compute the Dice loss~\cite{milletari2016v} between rasterized polygons as auxiliary loss:
\begin{equation}
\mathcal{L}_{\text{ras}}^{m} = \mathds{1}_{\left \{ m\leq M^\text{gt} \right \}} \text{Dice}\big(R(P_{m}), R(\hat{P}_{\hat{\sigma}(m)})\big)
\end{equation}
where $R$ indicates the rasterized mask of a given polygon, using a differentiable rasterizer~\cite{lazarow2022instance}.
We only compute $\mathcal{L}_\text{coord}^{m}$ and $\mathcal{L}_\text{ras}^{m}$ for predicted polygons with matched non-padded \gt{} while computing $\mathcal{L}_\text{cls}^{m}$ for all predicted polygons (including $\varnothing$).
The total loss $\mathcal{L}$ is then defined as:
\begin{equation}\label{eq:loss}
\mathcal{L}= \sum_{m}^{M}(\lambda_{\text{cls}} \mathcal{L}_{\text{cls}}^{m} + \lambda_{\text{coord}} \mathcal{L}_{\text{coord}}^{m}+\lambda_{\text{ras}} \mathcal{L}_{\text{ras}}^{m})
\end{equation}

\subsection{Towards Semantically-Rich Floorplans}\label{sec:functionality}

Our method can easily be extended to classify different room types and reconstruct additional architectural details such as doors and windows, while using the same input.
\parag{Room types.} 
The two-level polygon queries make our pipeline very flexible to be extended to identify room types.
We denote the output embedding from the last layer of the Transformer decoder as $D^{last}\in \mathbb{R}^{M\times N\times C}$.
We then aggregate room-level features by simply averaging corner-level features and obtain an aggregated embedding $\widehat{D}^{last}\in \mathbb{R}^{M\times C}$.
Finally, a simple linear projection layer predicts the room label probabilities using a softmax function.
Since $M$ is usually larger than the actual number of rooms in a scene, an additional empty class label is used to represent invalid rooms.
We denote $t_{m}$ as the type for polygon instance $V_{m}$.
We use the same matching permutation $\hat{\sigma}$ from Eq.\,\ref{eq:matcher} to find the matched prediction $\hat{t}_{\hat{\sigma}(m)}$.
The room type is supervised by a cross-entropy loss $\mathcal{L}_{{\text{room}\_\text{cls}}}^{m}$.

\parag{Doors and windows.}  A door or a window can be regarded as a line in a 2D \fp{}. 
Intuitively, we can view them as a special ``polygon" with 2 vertices.
This way, our pipeline can be directly adapted to predict doors and windows \emph{without} architecture modification.
 It is only required to increase the room-level queries $M$ since more polygons need to be predicted (Fig.\,\ref{fig:var1}).
Alternatively, we could use a separate line decoder to predict doors and windows.
Since a line can be parameterized by a fixed number of 2 vertices, we can either represent them as two-level queries but with a fixed number of corner-level queries (Fig.\,\ref{fig:var2}), or single-level queries (Fig.\,\ref{fig:var3}). The two-level queries variant is simply an adaptation of our polygon decoder.
For the single-level queries variant, we follow LETR~\cite{xu2021line} that directly predicts the two endpoints from the query.
The performance of each variant is analyzed in Sec.\,\ref{sec:result_sem_rich}.

\begin{table*}[ht]
\centering
\setlength{\tabcolsep}{5pt}
\resizebox{\textwidth}{!}{
\begin{tabular}{lccccccccccccc}
\toprule
\cmidrule{1-14} 
& & & & &
\multicolumn{3}{c}{Room} & \multicolumn{3}{c}{Corner} & \multicolumn{3}{c}{Angle}\\
\cmidrule(r){6-8}	\cmidrule(r){9-11} \cmidrule(r){12-14}
Method & Venue & Fully-neural & Single-stage & t (s) & Prec. & Rec. & F1 & Prec. & Rec. & F1 & Prec. & Rec. & F1\\
\midrule
Floor-SP \cite{Chen19ICCV}& ICCV19 & \xmark & \xmark & 785 & 89.\,\,\, & 88.\,\,\, & 88.\,\,\, &  81.\,\,\, & 73.\,\,\, & 76.\,\,\, & 80.\,\,\, & 72.\,\,\, & 75.\,\,\,\\
MonteFloor \cite{stekovic2021montefloor}& ICCV21 & \xmark & \xmark & 71 & 95.6 & 94.4 & 95.0 &  88.5  & 77.2 & 82.5 & \textbf{86.3} & 75.4 & 80.5\\ 
\arrayrulecolor{black!10}\midrule\arrayrulecolor{black}
HAWP \cite{xue2020holistically} & CVPR20 & \cmark & \xmark & 0.02 & 77.7 & 87.6 & 82.3 &  65.8 & 77.0 & 70.9 & 59.9 & 69.7 & 64.4\\
LETR \cite{xu2021line} & CVPR21 & \cmark & \xmark & 0.04 & 94.5 & 90.0 & 92.2 &  79.7 & 78.2 & 78.9 & 72.5 & 71.3 & 71.9 \\		
HEAT \cite{Chen22CVPR}& CVPR22 & \cmark & \xmark & 0.11 & 96.9 & 94.0 & 95.4 &  81.7 & 83.2 & 82.5 & 77.6 & 79.0 & 78.3\\
\arrayrulecolor{black!10}\midrule\arrayrulecolor{black}
\name{} (Ours) & - & \cmark & \cmark & \textbf{0.01} & \textbf{97.9} & \textbf{96.7} & \textbf{97.3}  & \textbf{89.1} &  \textbf{85.3} & \textbf{87.2} & 83.0 & \textbf{79.5} & \textbf{81.2}\\
\bottomrule
\end{tabular}
}
\vspace{-4px}
\caption{
\textbf{Foorplan reconstruction scores on Structured3D test set~\cite{zheng2020structured3d}.}
Our method offers state-of-the-art results while being significantly faster than existing works.
Runtime is averaged over the test set.
Scores of prior works are as taken from \cite{Chen22CVPR, stekovic2021montefloor}. 
}
\vspace{-10px}
\label{tab:s3d}
\end{table*}

\section{Experiments}
\label{sec:experiments}
\subsection{Datasets and Metrics}

\parag{Datasets.}
Structured3D~\cite{zheng2020structured3d} is a large-scale photo-realistic dataset containing 3500 houses with diverse floor plans covering both Manhattan and non-Manhattan layouts. 
It contains semantically-rich annotations including doors and windows, and 16 room types.
We adhere to the pre-defined split of 3000 training samples, 250 validation samples and 250 test samples.
As in \cite{stekovic2021montefloor,Chen22CVPR},
we convert the registered multi-view RGB-D panoramas to point clouds, and project the point clouds along the vertical axis into density images of size 256$\times $256 pixels. 
The density value at each pixel is the number of projected points to that pixel divided by the maximum point number so that it is normalized to [0, 1].

SceneCAD~\cite{avetisyan2020scenecad} contains 3D room layout annotations on real-world RGB-D scans of ScanNet~\cite{dai2017scannet}.
We convert the layout annotations to 2D floorplan polygons.
Annotations are only available for the ScanNet training and validation splits, so
we train on the training split and report scores on the validation split.
We use the same procedure as in Structured3D to project RGB-D scans to density maps. 

\parag{Metrics.}
Following \cite{stekovic2021montefloor, Chen22CVPR}, for each \gt{} room, we loop through the predictions and find the best-matching reconstructed room in terms of IoU. For the matched rooms we then report precision, recall and F1 scores at three geometry levels: rooms, corners and angles.
We compute precision, recall and F1 scores also for the semantic enrichment predictions.
For the room type, the metrics are computed like the room metric described above, with the additional constraint that the predicted semantic label must match the \gt{}.
A  window or door is considered correct if its $L_{2}$ distance to the \gt{} element is \textless10 pixels.

\subsection{Implementation Details}
\parag{Model settings.}
We use the same ResNet-50 backbone as HEAT~\cite{Chen22CVPR}.
We generate multi-scale feature maps from the last three backbone stages without FPN.
The fourth scale feature map is obtained via a $3\times 3$ stride 2 convolution on the final stage.
All feature maps are reduced to 256 channels by a 1$\times$1 convolution.
The Transformer consists of 6 encoder and 6 decoder layers with 256 channels.
We use 8 heads and $N_{s}$=$4$ sampling points for the deformable attention module.
The number of room-level queries and corner-level queries is set to $M=20$ and $N=40$.

\parag{Training.} 
We use the Adam optimizer~\cite{kingma2014adam} with
a weight decay factor 1e-4. Depending on the dataset size, we train the model on Structured3D for 500 epochs with an initial learning rate 2e-4 and on SceneCAD for 400 epochs with an initial learning rate 5e-5. The learning rate decays by a factor of 0.1 for the last 20\% epochs.
We set the coefficients for the matcher and losses to $\lambda_{\text{cls}}$\,=\,$2$, $\lambda_{\text{coord}}$\,=\,$5$, $\lambda_{\text{ras}}$\,=\,$1$
and use a single TITAN RTX GPU with 24GB memory for training.

\subsection{Comparison with State-of-the-art Methods}

\begin{table}[!t]
\centering
\setlength{\tabcolsep}{4pt}
\resizebox{\columnwidth}{!}{
		\begin{tabular}{lcccccccc}
			\toprule
			\cmidrule{1-9} 
		&	 & Room &
\multicolumn{3}{c}{Corner} & \multicolumn{3}{c}{Angle}\\
            \cmidrule(r){3-3}
			\cmidrule(r){4-6}
			\cmidrule(r){7-9}
 			Method & t(s) & IoU & Prec. & Rec. & F1 & Prec. & Rec. & F1 \\
			\midrule
 			Floor-SP \cite{Chen19ICCV} & 26 & 91.6 & 89.4 & \textbf{85.8} & 87.6 & 74.3 & 71.9 & 73.1\\ 
			HEAT~\cite{Chen22CVPR}  & 0.12 & 84.9 & 87.8 & 79.1 & 83.2 & 73.2 & 67.8 & 70.4\\  %
			\name{} (Ours)                    & \textbf{0.01} & \textbf{91.7} & \textbf{92.5} & 85.3 & \textbf{88.8} & \textbf{78.0} & \textbf{73.7} & \textbf{75.8}\\   %
		  \bottomrule
		\end{tabular}
	}
\vspace{-4px}
\caption{\textbf{Foorplan reconstruction on the SceneCAD val\,set~\cite{avetisyan2020scenecad}.}
}
\vspace{-20px}
\label{tab:sceneCAD}
\end{table}

Results are summarized in Tab.\,\ref{tab:s3d} for Structured3D and in Tab.\,\ref{tab:sceneCAD} 
 for SceneCAD.
We compare our method to a range of prior approaches that can be grouped into two broad categories:
Floor-SP\,\cite{Chen19ICCV} and MonteFloor\,\cite{stekovic2021montefloor} both rely on Mask R-CNN to segment rooms, followed by learning-free optimization techniques to recover the floorplan.
HAWP~\cite{xue2020holistically} and LETR~\cite{xu2021line} are originally generic methods to detect line segments and have been adapted to floorplan reconstruction in~\cite{Chen22CVPR}. 
HEAT \cite{Chen22CVPR} is an end-to-end trainable neural model that first detects corners, then links them via edge classification. 
Our \name{} outperforms all previous methods on Structured3D (Tab.\,\ref{tab:s3d}), increasing the F1 score by +$1.9$ for rooms, +$4.7$ for corners and +$2.9$ for angles from the previous state-of-the-art, HEAT. 
Our \name{} is the fastest one among the tested methods, with more than 10 times faster inference than HEAT.
MonteFloor additionally employs a Douglas-Peucker algorithm~\cite{douglas1973algorithms} for post-processing to simplify the topology of the output polygons. By contrast, \name{} does 
not rely on any post-processing steps.

On SceneCAD~(Tab.\,\ref{tab:sceneCAD}), we compare with two representative methods (for which code is available) from optimization-based and fully-neural categories Floor-SP and HEAT.
\name{} achieves notable improvement over other methods, especially on corner/angle precision.
For more details, please see the supplementary.
\parag{Cross-data generalization}.
We further evaluate the ability of our model to generalize across datasets.
For that, we train on Structured3D training set and evaluate on SceneCAD validation set (without fine-tuning on SceneCAD).
We compare with the current state-of-the-art, end-to-end method HEAT and report scores in Tab.\,\ref{tab:sceneCAD_generalize}.
Our model generalizes better to unseen data characteristics.
It outperforms HEAT significantly in almost every metric and particularly on room IoU~
(74.0 \vs{} 52.5).
We attribute the better generalization to the learned global reasoning capacity rather than focusing on separate corner detection and edge classifications as in HEAT.

\begin{table}[t]
\centering
\setlength{\tabcolsep}{4pt}
\resizebox{\columnwidth}{!}{
		\begin{tabular}{lcccccccc}
			\toprule
			\cmidrule{1-9} 
		&	 & Room &
\multicolumn{3}{c}{Corner} & \multicolumn{3}{c}{Angle}\\
            \cmidrule(r){3-3}
			\cmidrule(r){4-6}
			\cmidrule(r){7-9}
 			Method & t(s) & IoU & Prec. & Rec. & F1 & Prec. & Rec. & F1 \\
			\midrule
 			HEAT~\cite{Chen22CVPR} & 0.12 & 52.5 & 50.9 & 51.1 & 51.0 & 42.2 &  42.0 & 41.6\\  %
			\name{} (Ours) & \textbf{0.01} & \textbf{74.0} & \textbf{56.2} & \textbf{65.0} & \textbf{60.3} & \textbf{44.2} & \textbf{48.4} & \textbf{46.2}\\  %
		  \bottomrule
		\end{tabular}
	}
\vspace{-4px}
\caption{\textbf{Cross-data generalization.}
Models are trained on Structured3D train set but evaluated on SceneCAD val set. Our method shows significant robustness when the train-test domains differ.
}
\vspace{-10px}
\label{tab:sceneCAD_generalize}
\end{table}

\begin{table}[!t]
\centering
\setlength{\tabcolsep}{8pt}
\resizebox{\columnwidth}{!}{
		\begin{tabular}{lccccc}
			\toprule
			\cmidrule{1-6} 
				& Door/Window & Room$^{\ast }$ & Room & Corner & Angle \\ 
				\cmidrule(r){2-2} \cmidrule(r){3-3}
				\cmidrule(r){4-4}
				\cmidrule(r){5-5}
				\cmidrule(r){6-6}
			Method &  F1  & F1  & F1  & F1  & F1
\\

\midrule
			SD-TQ~(Fig.\,\ref{fig:var1}) &  81.1 & 70.7 & 94.3 & 83.9 & 76.7  \\    
            TD-TQ~(Fig.\,\ref{fig:var2}) & 80.8 & 71.4 & 93.4 & 82.0 & 73.7  \\ 
			TD-SQ~(Fig.\,\ref{fig:var3}) & 81.7 & 74.4 & 94.9 & 84.2 &  75.9 \\ 
			\bottomrule
		\end{tabular}
            }
\vspace{-4px}
\caption{\textbf{Semantically-rich floorplan reconstruction scores on Structured3D test set.}
The Room$^{\ast}$ metric is similar to Room, but additionally considers the correct room type classification.
}
\vspace{-20px}
\label{tab:semantic_rich}
\end{table}

\begin{figure}[!t]
    \centering \includegraphics[width=0.48\textwidth]{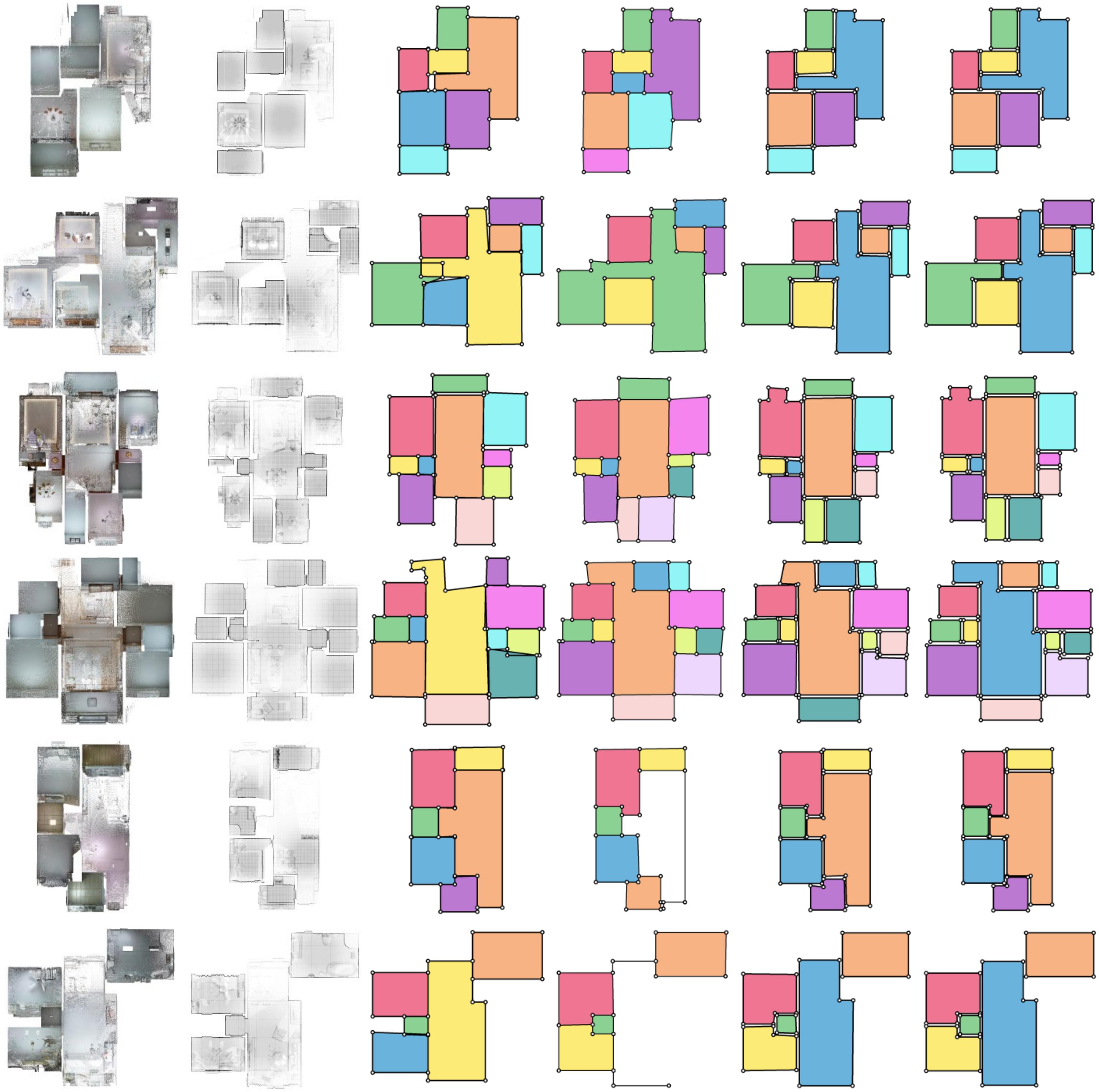}
    
    {\scriptsize 3D Scan} \hspace{0.2cm}
    {\scriptsize Density Map}
    \hspace{0.00cm}
    {\scriptsize   MonteFloor~\cite{stekovic2021montefloor}}
    \hspace{0.00cm}
    {\scriptsize   HEAT~\cite{Chen22CVPR}}
    \hspace{0.3cm}
    {\scriptsize \textbf{Ours}}
    \hspace{0.2cm}
    {\scriptsize Ground Truth}
    \caption{\textbf{Qualitative evaluations on Structured3D~\cite{zheng2020structured3d}.} Best viewed in color on a screen and zoom in. Colors are assigned based on room locations, \emph{without} semantic meaning.
    }
    \label{fig:s3d_result}
    \vspace{-10px}
\end{figure}

\begin{figure}[!t]
    \centering
    \includegraphics[width=0.48\textwidth]{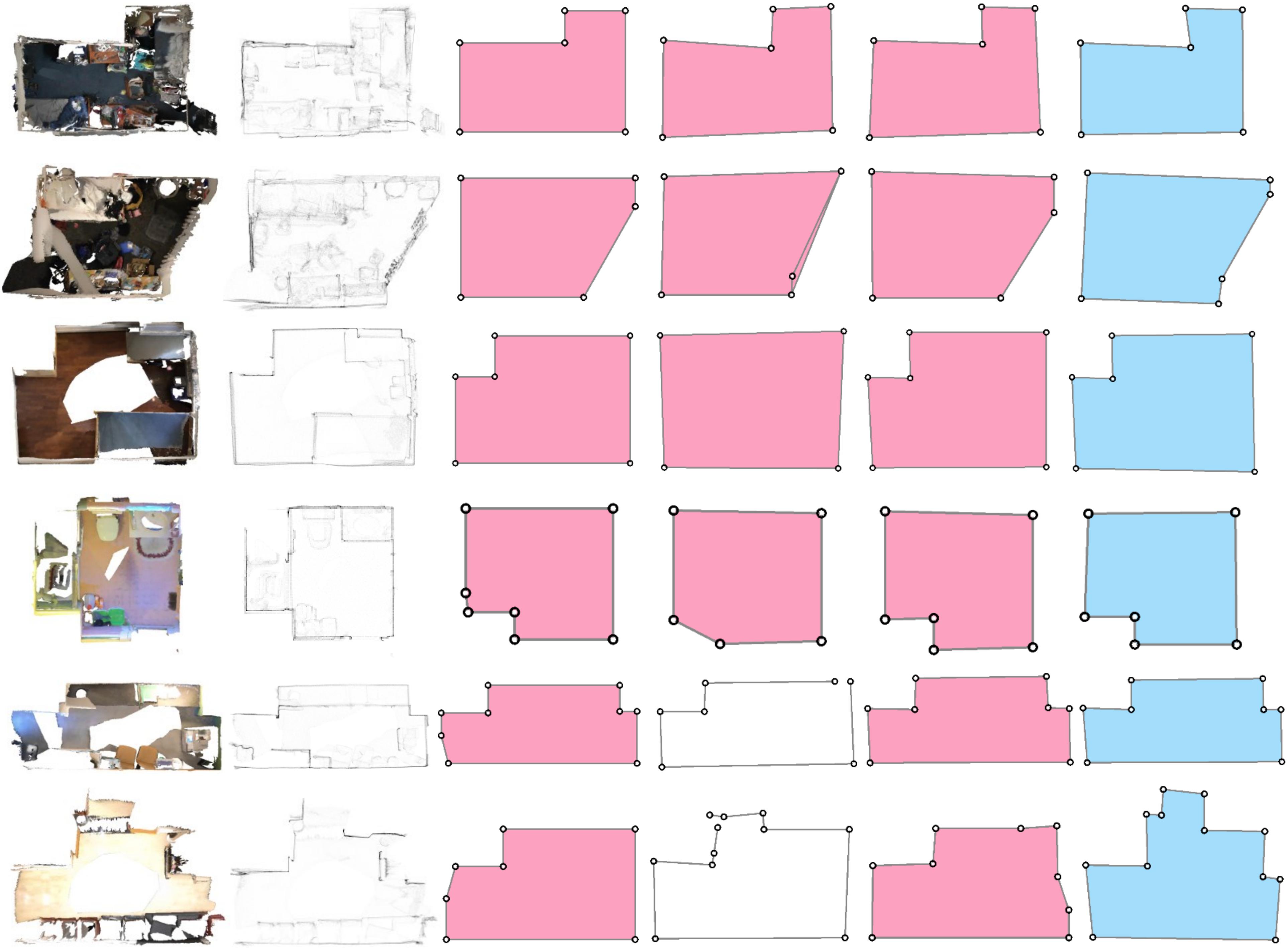}

    \hspace{0.1cm}
    {\scriptsize 3D Scan} 
    \hspace{0.1cm}
    {\scriptsize Density Map}
    \hspace{0.1cm}
    {\scriptsize   Floor-SP~\cite{Chen19ICCV}}
    \hspace{0.1cm}
    {\scriptsize   HEAT~\cite{Chen22CVPR}}
    \hspace{0.4cm}
    {\scriptsize \textbf{Ours}}
    \hspace{0.2cm}
    {\scriptsize Ground Truth}
    \caption{\textbf{Qualitative evaluations on SceneCAD~\cite{avetisyan2020scenecad}.} 
     HEAT is affected by missing corners and edges~(last 4 rows). Floor-SP tends to produce redundant corners due to its containment constraint~(4\textsuperscript{th}, 5\textsuperscript{th} row). Our method is more robust in these cases.
    }
\label{fig:scenecad_result}
\vspace{-16px}
\end{figure}

\parag{Qualitative results} on Structured3D are shown in Fig.\,\ref{fig:s3d_result}. 
The quality of \fp{} generated by two-stage pipelines is strongly affected by errors in the first stage, \eg, missing rooms with MonteFloor (3\textsuperscript{rd}, 4\textsuperscript{th} row) and missing corners/edges with HEAT (5\textsuperscript{th}, 6\textsuperscript{th} row). 
Instead, our holistic single-stage model produces more accurate predictions while being able to capture geometric details. 
We observe a similar pattern in Fig.\,\ref{fig:scenecad_result}.
HEAT suffers from missing corners/edges when the input point cloud is sparse (last 4 rows), while our \name{} handles these cases more robustly. Floor-SP forces the generated polygon to completely contain its room segmentation mask which, however, results in redundant corners (4\textsuperscript{th}, 5\textsuperscript{th} row). By contrast, \name{} produces more plausible results without imposing any hard constraints.

\subsection{Semantically-Rich \Fp{}s}\label{sec:result_sem_rich}
\begin{figure}
    \centering
    \includegraphics[width=0.47\textwidth]{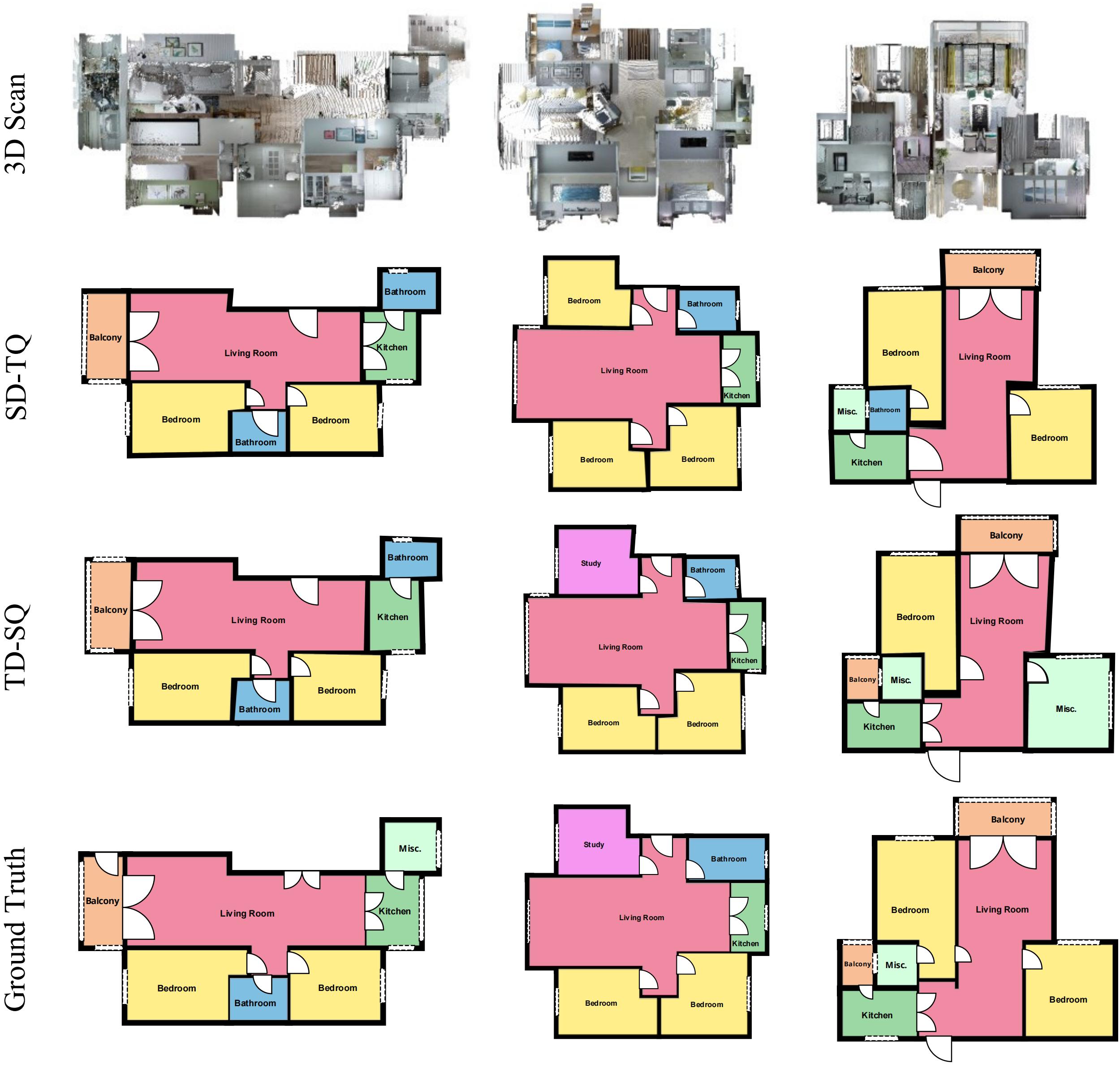}
\vspace{-4px}
    \caption{\textbf{Results of semantic-rich floorplans.}
    Dashed lines represent windows, arcs represent doors.
    The radius, number and orientation of arcs are determined by the length and direction of the predicted lines.
    Best viewed in color on a screen. 
    }
    \label{fig:windowdoor}
    \vspace{-15px}
\end{figure}

The quantitative results on semantically-rich \fp{} are summarized in Tab.\,\ref{tab:semantic_rich}. 
We observe that separating the room and door/window decoding can help improve room type classification since room and door/window may have significantly different geometry and semantic properties. However, the second-best performing method is SD-TQ where we use our original model with more polygon queries to model the doors and windows. For the two variants with line decoder, the single-level queries~(TD-SQ) work better than the two-level queries~(TD-TQ), suggesting that for shapes represented by a fixed number of parameters~(\eg{}, lines), single-level queries are sufficient.
We show our qualitative results on rich \fp{} reconstruction in Fig.\,\ref{fig:windowdoor}. For visualization purposes, we use arcs to represent doors.
Interestingly, our single decoder variant (SD-TQ) incorrectly identifies a room as ``bathroom" instead of ``misc''~(3\textsuperscript{rd} col.). However, this follows house design principles better than the \gt{}, since each house usually has a bathroom.

\subsection{Analysis Experiments}
\label{sec:ablation}
\parag{Two-level \vs{} single-level queries.} We model \fps{} as two-level queries. To validate that choice we compare with single-level queries, where a single query is responsible for predicting all ordered vertices of a room. Tab.\,\ref{tab:ablition_query}  shows that two-level queries greatly improve all metrics. The reason is that the two-level design relaxes the role of each query to model a single vertex rather than a sequence. Furthermore, it enables explicit interactions between vertices of the same room, and vertices across adjacent rooms, while single-level queries only enable room-level interactions.

\parag{Multi-scale feature maps.}
We leverage multi-scale feature maps to aggregate both local and global contexts for joint vertex positioning and order capturing. To validate this design, we conduct ablations by using only a single-level feature map obtained from the last stage of ResNet-50. 
Tab.\,\ref{tab:ablition_ms_refine} shows that multi-scale feature maps significantly improve all metrics, which indicates that local and global contexts are crucial for our Transformers for structured reasoning.

\parag{Iterative polygon refinement.}
We propose to directly learn vertex sequence coordinates as queries, refine them iteratively layer-by-layer, and use the updated positions as new reference points for deformable cross-attention. We ablate this by removing the refinement process and keeping the reference points static in intermediate layers while only updating the decoder embeddings.
Tab.\,\ref{tab:ablition_ms_refine} suggests that the refinement strategy significantly improves the performance. 

\parag{Loss functions.}
We use three loss components to supervise our network (Tab.\,\ref{tab:ablition_loss}).
Since the vertex label classification loss is essential and cannot be removed, we ablate for $\mathcal{L}_{\text{coord}}$ and $\mathcal{L}_{\text{ras}}$.
In the first experiment, we remove $\mathcal{L}_{\text{coord}}$ and replace the matching cost for sequence coordinates with a matching cost for the rasterized mask.
This leads to a significant drop in all metrics.
Next, we only remove $\mathcal{L}_{\text{ras}}$ which incurs a smaller drop in all metrics.
We conclude that the sequence coordinates regression loss is essential and the rasterization loss serves as an auxiliary loss.
\begin{table}[!t]
\setlength{\tabcolsep}{8pt}
\resizebox{\columnwidth}{!}{
\begin{tabular}{lllllll}
\toprule
\cmidrule{1-7} 
&
\multicolumn{2}{c}{Room} & \multicolumn{2}{c}{Corner} & \multicolumn{2}{c}{Angle}\\
\cmidrule(r){2-3}	\cmidrule(r){4-5} \cmidrule(r){6-7}
Queries  & Prec. & Rec. & Prec. & Rec. & Prec. & Rec. \\
\midrule
Single-level  & 74.4 & 73.4 & 65.1 & 58.9 & 61.4 & 55.6
\\
Two-level (Ours)  & \textbf{96.5} & \textbf{95.3} & \textbf{91.2} & \textbf{82.8}  & \textbf{88.3} & \textbf{80.3}\\ 
\bottomrule
\end{tabular}
}
\vspace{-6px}
\caption{
\textbf{Query analysis.}
Comparison between two-level and single-level queries.
Scores are on Structured3D validation set.}
\label{tab:ablition_query}
\vspace{-10px}
\end{table}

\begin{table}[!t]
\centering
\setlength{\tabcolsep}{6pt}
\resizebox{\columnwidth}{!}{
\begin{tabular}{cccccccc}
\toprule
\cmidrule{1-8} 
Multi-Scale & Polygon &
\multicolumn{2}{c}{Room} & \multicolumn{2}{c}{Corner} & \multicolumn{2}{c}{Angle}\\ %
\cmidrule(r){3-4}	\cmidrule(r){5-6} \cmidrule(r){7-8}
Features & Refinement & Prec. & Rec. & Prec. & Rec. & Prec. & Rec. \\
\midrule
 - &  \cmark & 93.9 & 92.9 & 87.8 & 79.6  & 83.3 & 75.6\\
\cmark & - &  94.8 & 93.0 & 88.7 &  80.7 & 84.2 & 76.7\\
\cmark  & \cmark & \textbf{96.5} & \textbf{95.3} & \textbf{91.2} & \textbf{82.8} & \textbf{88.3} & \textbf{80.3}  \\
\bottomrule
\end{tabular}
}
\vspace{-6px}
\caption{
\textbf{Model analysis.}
Impact of multi-scale features and polygon refinement.
Scores are on Structured3D validation set.}
\label{tab:ablition_ms_refine}
\vspace{-10px}
\end{table}

\begin{table}[!t]
\setlength{\tabcolsep}{6pt}
\resizebox{\columnwidth}{!}{
\centering
\scalebox{0.65}{
\renewcommand\arraystretch{1}
\begin{tabular}{ccccccccc}
\toprule
\cmidrule{1-9} 
\multicolumn{3}{c}{Settings} &
\multicolumn{2}{c}{Room} & \multicolumn{2}{c}{Corner} & \multicolumn{2}{c}{Angle}\\ \cmidrule(r){1-3}
\cmidrule(r){4-5}	\cmidrule(r){6-7} \cmidrule(r){8-9}
$\mathcal{L}_{\mathrm{cls}}$  &
$\mathcal{L}_{\mathrm{coord}}$ & $\mathcal{L}_{\mathrm{ras}}$ & Prec. & Rec. &  Prec. & Rec. &   Prec. & Rec. \\
\midrule
\cmark & - & \cmark & 84.3 & 83.2 & 75.4 &  68.8 & 70.6 & 64.5 \\
\cmark & \cmark & - & 96.0 & 94.5 & 89.9 &  81.7 & 87.0 & 79.2 \\
\cmark & \cmark & \cmark & \textbf{96.5}  & \textbf{95.3} & \textbf{91.2} &  \textbf{82.8} & \textbf{88.3} & \textbf{80.3}\\
\bottomrule
\end{tabular}
}
}
\vspace{-6px}
\caption{\textbf{Loss analysis.}
Impact of various losses.
$\mathcal{L}_{\text{cls}}$ is always required for determining valid corners.
The combination of $\mathcal{L}_{\text{coord}}$ and $\mathcal{L}_{\text{ras}}$ yields best results.
Scores are on Structured3D val set.}
\label{tab:ablition_loss}
\vspace{-15px}
\end{table}

\section{Conclusion}
\label{sec:conclusion}
In this work, we have introduced \name{}, a simple and direct model for 2D \fp{} reconstruction formulated as a polygon estimation problem.
The network learns to predict a varying number of rooms per \fp{}, each room represented as a varying length of ordered corner sequence.
Our single-stage, end-to-end trainable model shows significant improvements over prior multi-stage and heuristics-driven methods, both in performance and speed metrics.
Moreover, it can be flexibly extended to reconstruct semantically-rich floorplans.
We hope our approach inspires more applications in polygonal reconstruction tasks.

{\small\parag{Acknowledgments} We thank the authors of HEAT and MonteFloor for providing results on Structured3D for better comparison. Theodora Kontogianni and Francis Engelmann are postdoctoral research fellows at the ETH AI Center.}

\balance

{\small
\bibliographystyle{ieee_fullname}
\bibliography{egbib}
}
\clearpage
\appendix
\begin{appendices}
 
In the appendix, we first provide additional ablation studies on the attention design and numbers of the two-level queries (Sec.\,\ref{app:ablation}). Then we describe an alternative `tight' room layout used by some baselines
and report the scores of our model trained with this layout (Sec.\,\ref{app:tight_room}).We also provide
further implementation details and more results on semantically-rich floorplans (Sec.\,\ref{app:semantic_rich}).
Finally, we provide details for running competing methods Floor-SP~\cite{Chen19ICCV} and HEAT~\cite{Chen22CVPR}, as well as a learning-free baseline on the SceneCAD dataset~\cite{avetisyan2020scenecad} (Sec.\,\ref{app:run_compete}).

\section{Additional Ablation Studies}\label{app:ablation}

\subsection{Self-attention of the two-level queries}

In the main paper, the Transformer decoder performs self-attention on all vertex-level queries regardless of the polygon they belong to (Tab.\,\ref{tab:ablition_atten_mask}, \circlenum{2}).
Alternatively, in this experiment, we restrict the vertex-level queries to attend only to vertices within the same polygon. 
Implementation-wise, we add an attention mask to prevent attention from vertex-level queries in one polygon to vertex-level queries of another polygon. 
 We find that this restricted form of attention leads to an overall reduced performance (Tab.\,\ref{tab:ablition_atten_mask},~\circlenum{1}). 
We conclude that the self-attention between vertices across all polygons plays an important role for structured reasoning.
In particular, the attention mechanism across multiple polygons seems to help fine-tune the vertex positions of one polygon by attending to the vertex positions of its neighboring polygons.

\begin{table}[ht]
\centering
\setlength{\tabcolsep}{4pt}
\resizebox{\columnwidth}{!}{
		\begin{tabular}{lccccccccc}
			\toprule
			\cmidrule{1-10} 
			 & \multicolumn{3}{c}{Room} &
\multicolumn{3}{c}{Corner} & \multicolumn{3}{c}{Angle}\\
            \cmidrule(r){2-4}
			\cmidrule(r){5-7}
			\cmidrule(r){8-10}
			Settings  & Prec. & Rec. & F1 & Prec. & Rec. & F1 & Prec. & Rec.  & F1  \\
			\midrule
			\circlenum{1} 

           Intra-Poly. Attn.  & 95.5 & 94.2 &  94.8 & 90.3 & 81.9 & 85.9 & 87.2 &  79.1 & 83.0\\
			\circlenum{2} Inter-Poly. Attn.  & \textbf{96.5} & \textbf{95.3} & \textbf{95.9} & \textbf{91.2} & \textbf{82.8} & \textbf{86.8} & \textbf{88.3} & \textbf{80.3} & \textbf{84.1}  \\

			\bottomrule
		\end{tabular}
	}
        \vspace{-4px}
	\caption{
	\textbf{Impact of self-attention design on Structured3D val.} Using self-attention only between the vertices of a single polygon leads to a drop in the F1 score,
 showing the usefulness of extending the effect of vertices across all polygons.}
	\label{tab:ablition_atten_mask}
\end{table}

\subsection{Number of queries}

We study the effect of different numbers of queries in each level in Tab.\,\ref{tab:ablition_num_query}. The number of queries is based on: 1) the maximum number of rooms and the maximum number of corners in each room in the training dataset. 2) computational efficiency. Although the number of queries empirically has a small impact on reconstruction quality, the results highlight the robustness of our model towards fewer queries and justify the associated gain
in runtime (16 ms → 8 ms), especially since we aim
to be much faster than prior work.
We chose $M=20$ and $N=40$ as a good compromise between model performance, inference time, as well as training time.

\begin{table}[ht]
\centering
\setlength{\tabcolsep}{8pt}
\resizebox{\columnwidth}{!}{
\begin{tabular}{ccccccccc}
\toprule
\cmidrule{1-9} 
\multicolumn{2}{l}{Settings} & &
\multicolumn{2}{c}{Room} & \multicolumn{2}{c}{Corner} & \multicolumn{2}{c}{Angle}\\ \cmidrule(r){1-2}
\cmidrule(r){4-5}	\cmidrule(r){6-7} \cmidrule(r){8-9}
$\npoly$ & $\nvert$ & t (ms) & Prec. & Rec. & Prec. & Rec. & Prec. & Rec. \\
\midrule
15 & 30 & 8.4  & 95.7 & 94.5 & 90.2 & 82.2 & 86.5 & 79.0 \\
20 & 40 & 10.8 & 96.3 & 95.0 & 90.8 & {82.7} & 87.8 & 80.0 \\
20 & 50 & 12.8 & 96.2 & 94.6 & 90.5 & 82.1 & 87.6 & 79.5 \\
30 & 40 & 13.9 & {96.8} & {95.5} & {91.1} & {82.7} & {88.1} & {80.1} \\
30 & 50 & 16.8 & 96.0 & 94.8 & 90.6 & 82.4 & 87.5 & 79.7 \\
\bottomrule
\end{tabular}
}
\vspace{-4px}
\caption{\textbf{Analysis on number of queries.} The reported scores are on Structured3D validation set averaged over three runs.}
\label{tab:ablition_num_query}
\end{table}

\section{Tight Room Layouts}
\label{app:tight_room}

\begin{table}[b]
\centering
\setlength{\tabcolsep}{4pt}
\resizebox{\columnwidth}{!}{
\begin{tabular}{lccccccccc}
\toprule
\cmidrule{1-10} 
&
\multicolumn{3}{c}{Room} & \multicolumn{3}{c}{Corner} & \multicolumn{3}{c}{Angle}\\
\cmidrule(r){2-4}	\cmidrule(r){5-7} \cmidrule(r){8-10}
Method  & Prec. & Rec. & F1 & Prec. & Rec. & F1 & Prec. & Rec. & F1\\
\midrule
Floor-SP \cite{Chen19ICCV} & 89.\,\,\, & 88.\,\,\, & 88.\,\,\, &  81.\,\,\, & 73.\,\,\, & 76.\,\,\, & 80.\,\,\, & 72.\,\,\, & 75.\,\,\,\\
MonteFloor \cite{stekovic2021montefloor}&  95.6 & 94.4 & 95.0 &  88.5  & 77.2 & 82.5 & \second{86.3} & 75.4 & 80.5\\ 
\arrayrulecolor{black!10}\midrule\arrayrulecolor{black}
HAWP \cite{xue2020holistically} &  77.7 & 87.6 & 82.3 &  65.8 & 77.0 & 70.9 & 59.9 & 69.7 & 64.4\\
LETR \cite{xu2021line} & 94.5 & 90.0 & 92.2 &  79.7 & 78.2 & 78.9 & 72.5 & 71.3 & 71.9 \\		
HEAT \cite{Chen22CVPR} & 96.9 & 94.0 & 95.4 &  81.7 & 83.2 & 82.5 & 77.6 & 79.0 & 78.3\\
\arrayrulecolor{black!10}\midrule\arrayrulecolor{black}
\name{} (Ours) & \first{97.9} & \first{96.7} & \first{97.3}  & \second{89.1} &  \first{85.3} & \second{87.2} & 83.0 & \second{79.5} & \second{81.2}\\
\name{}$^*$ (Ours) & \second{97.2} & \second{96.2} & \second{96.7}  & \first{91.6} &  \second{83.4} & \first{87.3} & \first{88.3} & \first{80.5} &  \first{84.2} \\
\bottomrule
\end{tabular}
}
\vspace{-4px}
\caption{
\textbf{Foorplan reconstruction scores on Structured3D test.}
For a complete and fair comparison, we complement Tab.\,\ref{tab:s3d} from the main paper with the result of our model trained on tight room layouts, indicated by $^*$. \first{Cyan} and \second{orange} mark the two top scores.}

\label{tab:s3d_tight}
\end{table}

\begin{figure*}[!t]
    \centering 
    \includegraphics[width=0.97\textwidth]{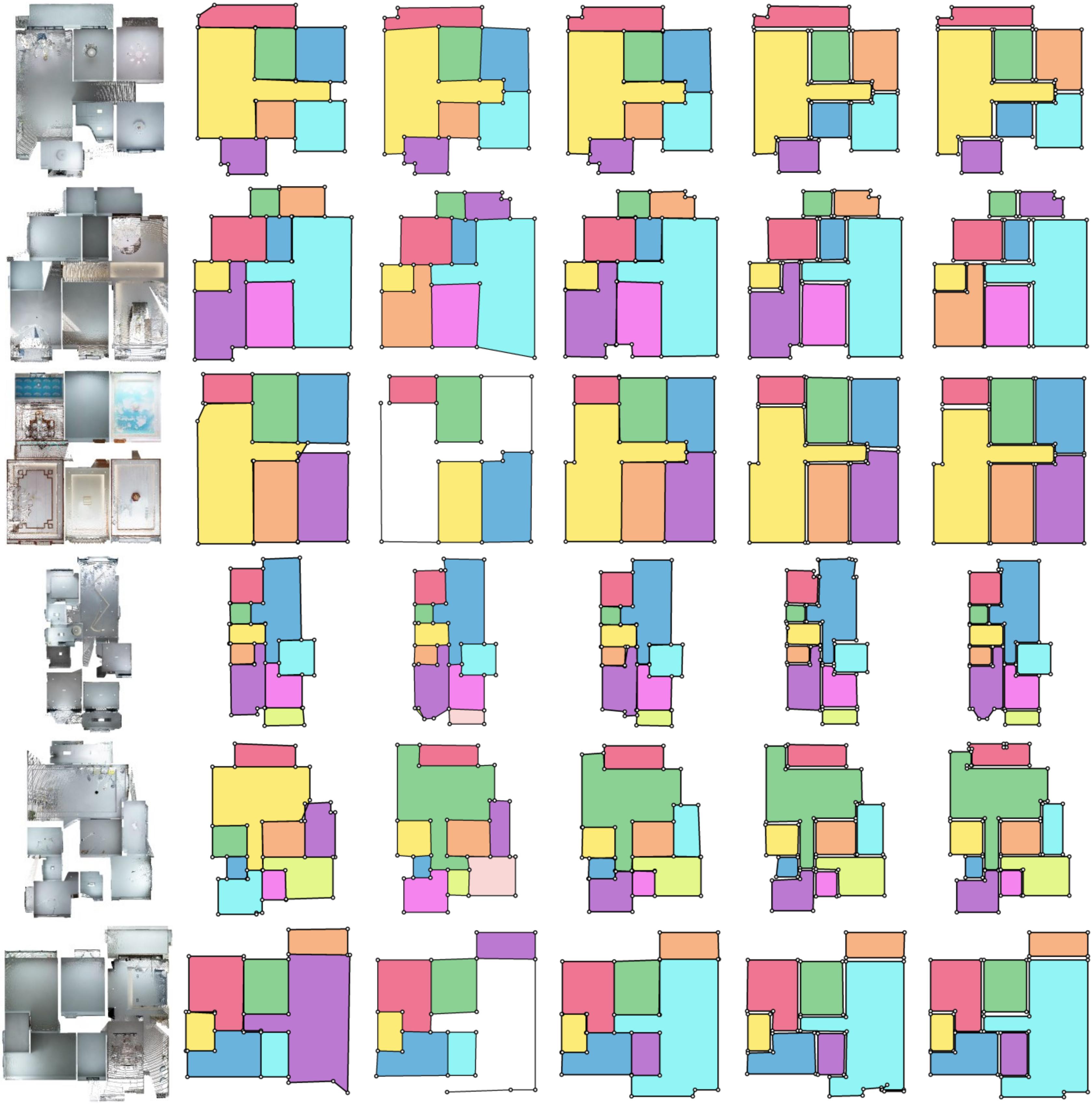}

    {3D Scan} 
    \hspace{1.2cm}
    {MonteFloor~\cite{stekovic2021montefloor}}
    \hspace{1cm}
    {HEAT~\cite{Chen22CVPR}}
    \hspace{1.6cm}
    {{Ours$^*$}}
    \hspace{1.6cm}
    {{Ours}}
    \hspace{1.3cm}
    {Ground Truth}
    
    \caption{\textbf{More qualitative evaluations on Structured3D~\cite{zheng2020structured3d}.} Colors are assigned based on room locations, \emph{without} semantic meaning. Ours$^*$ denotes the result of our model trained on tight room layout. (\textit{Best viewed in color on a screen and zoomed in.})
    }
    \label{fig:s3d_result_tight}
\end{figure*}

We represent floorplans as a set of closed polygons, which is consistent with the groundtruth annotations in Structured3D~\cite{zheng2020structured3d}.
The advantage of this representation is that the thickness of inner walls is implicitly provided by the distance between neighboring room polygons (see  Fig.\,\ref{fig:s3d_result_tight}, \emph{right column}).
Alternatively, HEAT~\cite{Chen22CVPR} predict floorplans as planar graphs, which means that the walls of adjacent rooms are represented by a single shared edge in the graph. In particular, this simpler graph representation can approximate the true floorplan only up to the thickness of the walls.
Furthermore, to train HEAT, an additional pre-processing step is required which merges the groundtruth edges of neighboring polygons into a single shared edge.
It is important to note that the evaluation still runs on the unmodified groundtruth floorplans.
Therefore, during evaluation, HEAT performs a post-processing step to obtain a set of closed room polygons from the estimated planar graph.

To show that the improved performance of our model is independent of the layout representation,
we train a model on the same `tight' room layout representation as HEAT and report the scores in Tab.\,\ref{tab:s3d_tight} marked with a star ($^*$).
Note again that all models are evaluated on the same non-processed groundtruth annotations of Structured3D~\cite{zheng2020structured3d}.
We observe that the room metrics of the model trained on tight room layouts drop a bit compared with our original model, while still outperforming all other methods.
The drop in scores is not surprising since the modified training data is only an approximation of the original groundtruth used for evaluation.
Furthermore, the room metrics penalize overlap between rooms~\cite{stekovic2021montefloor}. 
Results from tight room layouts are more likely to overlap, which can potentially degrade the room metrics. 
Interestingly, the angle metrics improve, especially when it comes to angle precision, which already outperforms MonteFloor. 
In the tight room layout, we encourage corners in adjacent walls to share the same location and have exactly complementary angles. 
This implicit constraint might help the model to reason about angle relationships,
thus benefiting the angle metrics.
More qualitative results can be found in Fig.\,\ref{fig:s3d_result_tight}.

\section{Semantically-Rich Floorplan Models}\label{app:semantic_rich}

\begin{table*}[ht]
\centering
\setlength{\tabcolsep}{8pt}
\resizebox{\textwidth}{!}{
\begin{tabular}{lccccccccccccccc}
\toprule
\cmidrule{1-16} 
& \multicolumn{3}{c}{Door/Window} &
\multicolumn{3}{c}{Room$^*$} &
\multicolumn{3}{c}{Room} & \multicolumn{3}{c}{Corner} & \multicolumn{3}{c}{Angle}\\
\cmidrule(r){2-4} \cmidrule(r){5-7} \cmidrule(r){8-10}	\cmidrule(r){11-13} \cmidrule(r){14-16}
Method & Prec. & Rec. & F1 & Prec. & Rec. & F1 & Prec. & Rec. & F1 & Prec. & Rec. & F1 & Prec. & Rec. & F1\\
\midrule
SD-TQ & 83.4 & 79.0 & 81.1 & 71.5 & 70.0 & 70.7 & 95.3 & 93.3 & 94.3 & \textbf{86.0} & 81.8 & 83.9 & \textbf{78.6} & \textbf{74.9} & \textbf{76.7}\\
TD-TQ & 82.6 & \textbf{79.1} & 80.8 & 71.9 & 70.9 & 71.4 & 94.0 & 92.8 & 93.4 & 84.2 & 80.0 & 82.0 & 75.6 & 71.9 & 73.7\\
TD-SQ & \textbf{85.6} & 78.2 & \textbf{81.7} & \textbf{74.8} & \textbf{74.0} & \textbf{74.4} & \textbf{95.4} & \textbf{94.4} & \textbf{94.9} & 85.8 & \textbf{82.6} & \textbf{84.2} & 77.3 & 74.5 & 75.9\\
\bottomrule
\end{tabular}
}
\vspace{-4px}
\caption{
\textbf{Detailed scores of semantically-rich floorplan reconstruction on
Structured3D test set~\cite{zheng2020structured3d}.}
 }
\vspace{-10px}
\label{tab:semantic_rich_detail}
\end{table*}

We proposed three model variants for reconstructing 
semantically-rich floorplans: (1)\,SD-TQ: single decoder with two-level queries. (2)\,TD-TQ: two decoders with
two-level queries. (3)\,TD-SQ: two decoders with single-level queries in the line decoder.
Here, we explain additional implementation details along with more results.
We also describe our heuristic for drawing the door arcs based on the width of the door segments.

\subsection{Implementation details}
We describe the implementation details of the three model variants to extend our \name{} architecture to predict room types, doors and windows (Fig.\,\ref{fig:model_variant_sem_rich}, main paper).
\parag{SD-TQ.}
Single decoder with two-level queries.
We take our original architecture and increase the number of room-level queries $M$ from 20 to 70 while keeping the number of corner-level queries $N$ unchanged. In addition, since there is no need to rasterize lines, we remove the rasterization loss $\mathcal{L}_{\text{ras}}$. We predict all the semantic types (room types, door or window) from the aggregated room-level features of the output embedding of the polygon decoder, as described in Sec.\,\ref{sec:functionality} of the main paper.
\parag{TD-TQ.} Two decoders with two-level queries. We add a separate line decoder with the same architecture as the polygon decoder. In the line decoder, we set the number of line-level queries to 50 and the number of corner-level queries to 2. We remove the rasterization
loss $\mathcal{L}_{\text{ras}}$ term for the line decoder~(still used in the polygon decoder). Room types are predicted from the aggregated room-level features of the output embedding of the polygon decoder, while door/window types are predicted from the aggregated line-level features of the output embedding of the line decoder.
\parag{TD-SQ.} Two decoders with single-level queries. We add a separate line decoder that takes single-level queries, and predicts the coordinates of the two endpoints of each line directly, similar to~\cite{xu2021line}. We set the number of single-level queries to 50. Room types are predicted as in TD-TQ. Door/window types are predicted based on the output embedding of the line decoder.

\subsection{Plotting doors}
To obtain floorplan illustrations that are closer to actual floorplans used by architects, we follow the typical notation and represent doors as arcs.
Note that this step is only for visualization purposes and is not part of the annotations in the training datasets. In particular, we cannot predict towards which side a door opens and if it is a double door or a single door. The visualization used in this paper is based on a heuristic which creates a double door when the predicted width of the door exceeds a certain threshold.
The exact heuristic is shown in Algorithm\,\ref{alg:plot_door}.

\begin{algorithm}
\caption{Algorithm for plotting doors as arcs}\label{alg:plot_door}
\begin{algorithmic}[1]
\Require a set of predicted lines $L = \left \{ l_{i} \right \}_{i=1}^{N^{l}}$, where $l_{i}=\left ( \left ( x_{1}^{i}, y_{1}^{i}\right ),\left ( x_{2}^{i}, y_{2}^{i} \right ) \right )$
\Ensure plotting of a set of arcs

\State calculate the median length $m$ of all the lines $L$
\For{$l_{i}$ in $L$}
    \If{$\left \| y_{2}^{i}-y_{1}^{i} \right \| > \left \|x_{2}^{i}-x_{1}^{i}\right \|$}
        \If{$y_{2}^{i}> y_{1}^{i}$}
            $e_{1}^{i}=(x_{1}^{i},y_{1}^{i})$ and $e_{2}^{i}=(x_{2}^{i},y_{2}^{i})$
            
        \Else \hspace{0.1cm} $e_{1}^{i}=(x_{2}^{i},y_{2}^{i})$ and $e_{2}^{i}=(x_{1}^{i},y_{1}^{i})$
        \EndIf
        \If{$\text{len}(l_{i}) < 1.5 \times m$}
        draw a quadrant centered at $e_{1}^{i}$ with a radius of len($l_{i}$) from $e_{2}^{i}$ clockwise
        \Else \hspace{0.1cm}
        draw two opposite quadrants centered at $e_{1}^{i}$ and $e_{2}^{i}$ with a radius of $\text{len}(l_{i})/2$ on the right of $\overrightarrow{e_{1}^{i}e_{2}^{i}}$
        \EndIf

    \Else
        \If{$x_{2}^{i}> x_{1}^{i}$}
            $e_{1}^{i}=(x_{2}^{i},y_{2}^{i})$ and $e_{2}^{i}=(x_{1}^{i},y_{1}^{i})$
            
        \Else \hspace{0.1cm} $e_{1}^{i}=(x_{1}^{i},y_{1}^{i})$ and $e_{2}^{i}=(x_{2}^{i},y_{2}^{i})$
        \EndIf
        \If{$\text{len}(l_{i}) < 1.5 \times m$}
        draw a quadrant centered at $e_{1}^{i}$ with a radius of len($l_{i}$) from $e_{2}^{i}$ counterclockwise
        \Else \hspace{0.1cm}
        draw two opposite quadrants centered at $e_{1}^{i}$ and $e_{2}^{i}$ with a radius of $\text{len}(l_{i})/2$ on the left of $\overrightarrow{e_{1}^{i}e_{2}^{i}}$
        \EndIf
    \EndIf
    
\EndFor

\end{algorithmic}
\end{algorithm}

\begin{figure*}[!t]
    \centering \includegraphics[width=0.99\textwidth]{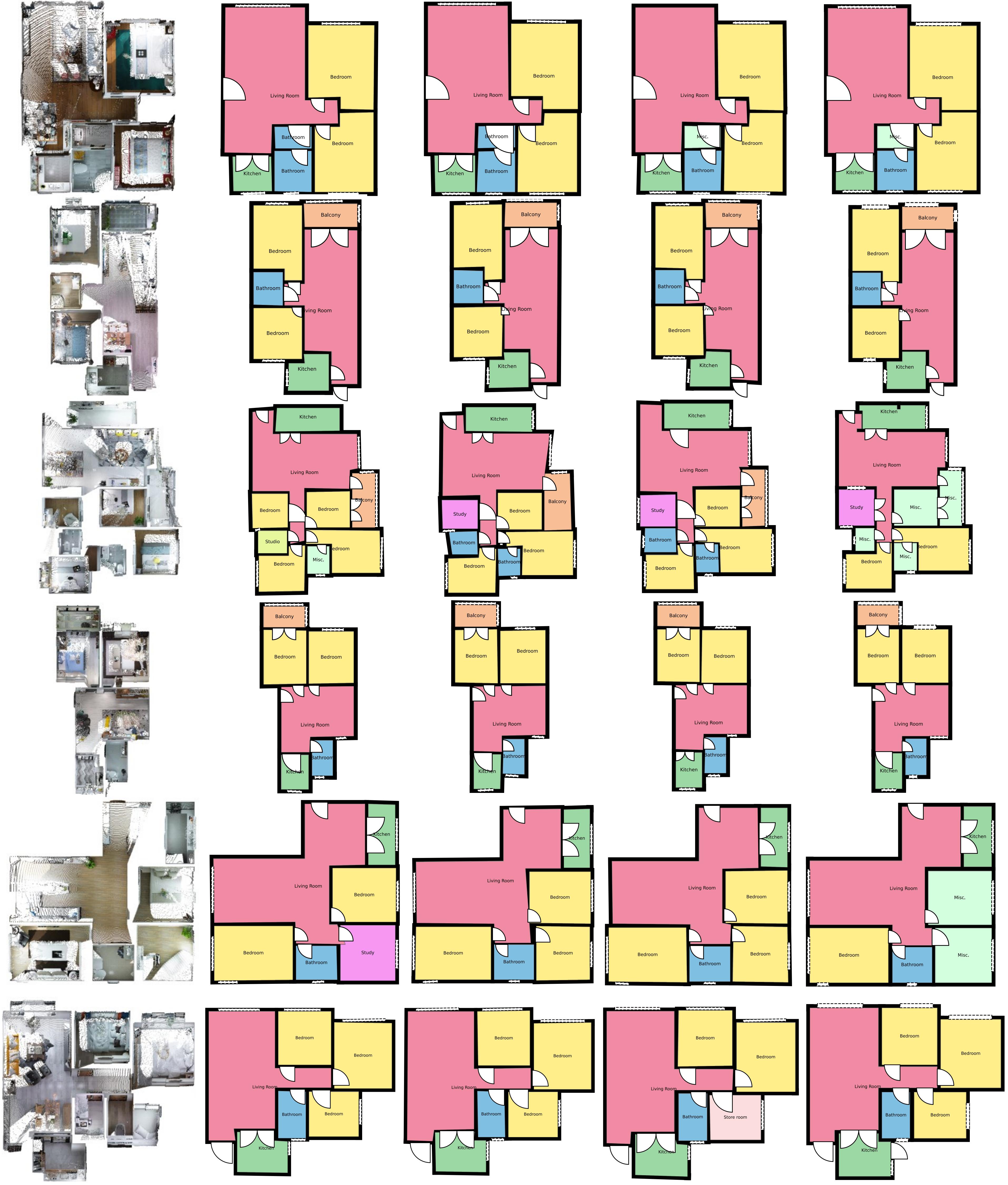}

    {3D Scan} 
    \hspace{2.2cm}
    {SD-TQ}
    \hspace{2.2cm}
    {TD-TQ}
    \hspace{2.2cm}
    {TD-SQ}
    \hspace{2.2cm}
    {Ground Truth}
    
    \caption{\textbf{Additional qualitative results on semantically-rich floorplans.} 
    (\textit{Best viewed in color on a screen and zoomed in.})
    }
    \label{fig:semantic_rich_more}
\end{figure*}

\subsection{Additional results} Tab.\,\ref{tab:semantic_rich_detail} complements  Tab.\,\ref{tab:semantic_rich} in the main paper by providing more detailed scores of the three model variants on semantically-rich floorplan reconstruction. By comparing the single-decoder variant (SD-TQ) with the two-decoder variants (TD-TQ and TD-SQ), we find that separate decoders can help improve room type classification. Our polygon queries are designed for geometries with a varying number of vertices. Since a line has a fixed number of 2 vertices, the single-level query variant (TD-SQ) works better than the two-level query variant (TD-TQ). We provide more qualitative results in Fig.\,\ref{fig:semantic_rich_more}.

\section{Running Competing Approaches}\label{app:run_compete}

For comparison on the SceneCAD dataset~\cite{avetisyan2020scenecad} in the main paper, we select two representative methods from both optimization-based and fully-neural categories Floor-SP~\cite{Chen19ICCV}
and HEAT~\cite{Chen22CVPR} that offer state-of-the-art results in floorplan reconstruction and have a public codebase. 
For a more complete comparison, here we also report the performance of a heuristic-guided pipeline that is free of any deep learning components.
Here we describe in detail how we adapted those methods for the SceneCAD dataset~\cite{avetisyan2020scenecad}.

\vspace{-10px}
\paragraph{Floor-SP:} We used the official implementation %
with a change in the sequential room-wise shortest path module. We project 3D scans into density images of size 256$\times $256 pixels. Unlike Structured3D, the SceneCAD dataset usually contains only one room per scene, which will result in larger occupancy pixel areas for a single room.
While this does not cause problems for HEAT and our approach, it can lead to a large search space for the sequential room-wise shortest path module in Floor-SP since each room mask contains a large number of pixels.
Using Floor-SP default settings, we observe the solver cannot find a solution for many scenes due to the computational complexity of the larger number of pixels per room.
Therefore, we down-sample the density map to 64$\times $64 pixels (a similar size with a single room in the density map of Structured3D) to help reduce the search space.
Please note, we train other modules (room mask extraction and corner/edge detection) on the original density map without down-sampling.
\vspace{-10px}
\paragraph{HEAT:} We used the official implementation %
with a batch size of 10. We train the model for 400 epochs and found that longer training times would not help to improve the performance further.
For the experiments on cross-data generalization, we directly load the released official checkpoints trained on Structured3D and evaluate them on SceneCAD.
\vspace{-10px}
\paragraph{Non-learned baseline:} We first project the 3D scan along the vertical axis into an occupancy map of size 256×256 pixels. A pixel is occupied if at least one point is projected to this pixel. To mitigate the impact of missing scans, we apply dilation and erosion to fill holes in the occupancy map. Then we employ a learning-free polygon vectorization algorithm~\cite{li2020approximating} to extract closed polygons from the occupancy map, finally followed by a Douglas-Peucker algorithm~\cite{douglas1973algorithms} to remove redundant corners.

\begin{table}[ht]
\centering
\setlength{\tabcolsep}{4pt}
\resizebox{\columnwidth}{!}{
		\begin{tabular}{lcccccccc}
			\toprule
			\cmidrule{1-9} 
		&	 & Room &
\multicolumn{3}{c}{Corner} & \multicolumn{3}{c}{Angle}\\
            \cmidrule(r){3-3}
			\cmidrule(r){4-6}
			\cmidrule(r){7-9}
 			Method & t(s) & IoU & Prec. & Rec. & F1 & Prec. & Rec. & F1 \\
			\midrule

            Non-learned  & 0.22 & 86.0 & 66.0 & 80.8 & 72.7 & 45.0 & 55.2 & 49.6 \\ 
    
 			Floor-SP \cite{Chen19ICCV} & 26 & 91.6 & 89.4 & \textbf{85.8} & 87.6 & 74.3 & 71.9 & 73.1\\ 
			HEAT~\cite{Chen22CVPR}  & 0.12 & 84.9 & 87.8 & 79.1 & 83.2 & 73.2 & 67.8 & 70.4\\  %
			\name{} (Ours)                    & \textbf{0.01} & \textbf{91.7} & \textbf{92.5} & 85.3 & \textbf{88.8} & \textbf{78.0} & \textbf{73.7} & \textbf{75.8}\\   %
		  \bottomrule
		\end{tabular}
	}
\vspace{-4px}
\caption{\textbf{Foorplan reconstruction on the SceneCAD val\,set~\cite{avetisyan2020scenecad}.}
}
\vspace{-10px}
\label{tab:sceneCAD_supp}
\end{table}

We complement Tab.\,\ref{tab:sceneCAD} of the main paper with the result of the non-learned baseline (the 1\textsuperscript{st} row in Tab.\,\ref{tab:sceneCAD_supp}). We report the running time of the polygon vectorization process. 
Surprisingly, the pipeline achieves a room IoU of 86.0 and a corner recall of 80.8. However, the angle metrics indicate that those polygons usually fail to accurately describe the geometry of the actual room shapes. This suggests that the non-learned baseline can only capture the rough shape of the floorplan compared with methods that utilize deep learning.

\end{appendices}
\end{document}